\documentclass[review]{elsarticle}

\usepackage{lineno,hyperref}
\usepackage{times}
\usepackage{epsfig}
\usepackage{graphicx}
\usepackage{amsmath}
\usepackage{amssymb}
\usepackage{mathrsfs}
\usepackage{algorithm}
\usepackage{helvet}
\usepackage{courier}
\usepackage{algpseudocode}
\usepackage{color}
\usepackage{multicol}
\usepackage{multirow}
\usepackage{array}
\usepackage{subfigure}
\usepackage{comment}
\usepackage{booktabs}
\usepackage{makecell}
\usepackage{array}
\usepackage{caption}
\usepackage{float}
\usepackage{xspace}

\makeatletter
\DeclareRobustCommand\onedot{\futurelet\@let@token\@onedot}
\def\@onedot{\ifx\@let@token.\else.\null\fi\xspace}

\def\eg{\emph{e.g}\onedot} 
\def\ie{\emph{i.e}\onedot}

\def\etal{\emph{et al}\onedot}
\makeatother

\journal{}

\begin{document}
	
	\begin{frontmatter}
		
		\title{From General to Specific: Online Updating for Blind Super-Resolution}
		
		\author[1,2]{Shang Li}
	\ead{lishang2018@ia.ac.cn}
	
	\author[1]{Guixuan Zhang}
	\ead{guixuan.zhang@ia.ac.cn}
	
	\author[1,2]{Zhengxiong Luo}
	\ead{luozhengxiong2018@ia.ac.cn}

	\author[1]{\\ Jie Liu\corref{cor1}}
	\ead{jie.liu@ia.ac.cn}
	
	\author[1]{Zhi Zeng}
	\ead{zhi.zeng@ia.ac.cn}
	
	\author[1,2]{Shuwu Zhang}
	\ead{shuwu.zhang@ia.ac.cn}
	
	\cortext[cor1]{Corresponding author}
	\address[1]{Institute of Automation, Chinese Academy of Sciences}
	\address[2]{School of Artificial Intelligence, University of Chinese Academy of Sciences\\ No.95, Zhongguancun East Road, Beijing 100190, PR China}

	\begin{abstract}
		Most deep learning-based super-resolution (SR) methods are not image-specific: 1) They are trained on samples synthesized by predefined degradations (\eg bicubic downsampling), regardless of the domain gap between training and testing data. 2) During testing, they super-resolve all images by the same set of model weights, ignoring the degradation variety. As a result, most previous methods may suffer a performance drop when the degradations of test images are unknown and various (\ie the case of blind SR). To address these issues, we propose an online SR (ONSR) method. It does not rely on predefined degradations and allows the model weights to be updated according to the degradation of the test image. Specifically, ONSR consists of two branches, namely internal branch (IB) and external branch (EB). IB could learn the specific degradation of the given test LR image, and EB could learn to super resolve images degraded by the learned degradation. In this way, ONSR could customize a specific model for each test image, and thus get more robust to various degradations. Extensive experiments on both synthesized and real-world images show that ONSR can generate more visually favorable SR results and achieve state-of-the-art performance in blind SR.
	\end{abstract}
	
	\begin{keyword}
		blind super-resolution \sep online updating \sep internal learning \sep external learning
	\end{keyword}
	
\end{frontmatter}

\section{Introduction}

Single image super-resolution (SISR) aims to reconstruct a plausible high-resolution (HR) image from its low-resolution (LR) counterpart. As a fundamental vision task, it has been widely applied in video enhancement, medical imaging and surveillance imaging. Mathematically, the HR image $\mathbf{x}$ and LR image $\mathbf{y}$ are related by a degradation model 
\begin{equation}\label{eq:1}
	\small
	\mathbf{y} = (\mathbf{k}\otimes \mathbf{x})\downarrow_{s} + \mathbf{n},
\end{equation} 
where $\otimes$ represents two-dimensional convolution of $\mathbf{x}$ with blur kernel $\mathbf{k}$, $\downarrow_{s}$ denotes the $s$-fold downsampler, and $\mathbf{n}$ is usually assumed to be additive, white Gaussian noise (AWGN) \cite{usr}. The goal of SISR is to restore the corresponding HR image of the given LR image, which is a classical ill-posed inverse problem. 

\begin{figure}[t]
	\centering
	\includegraphics[width=\linewidth]{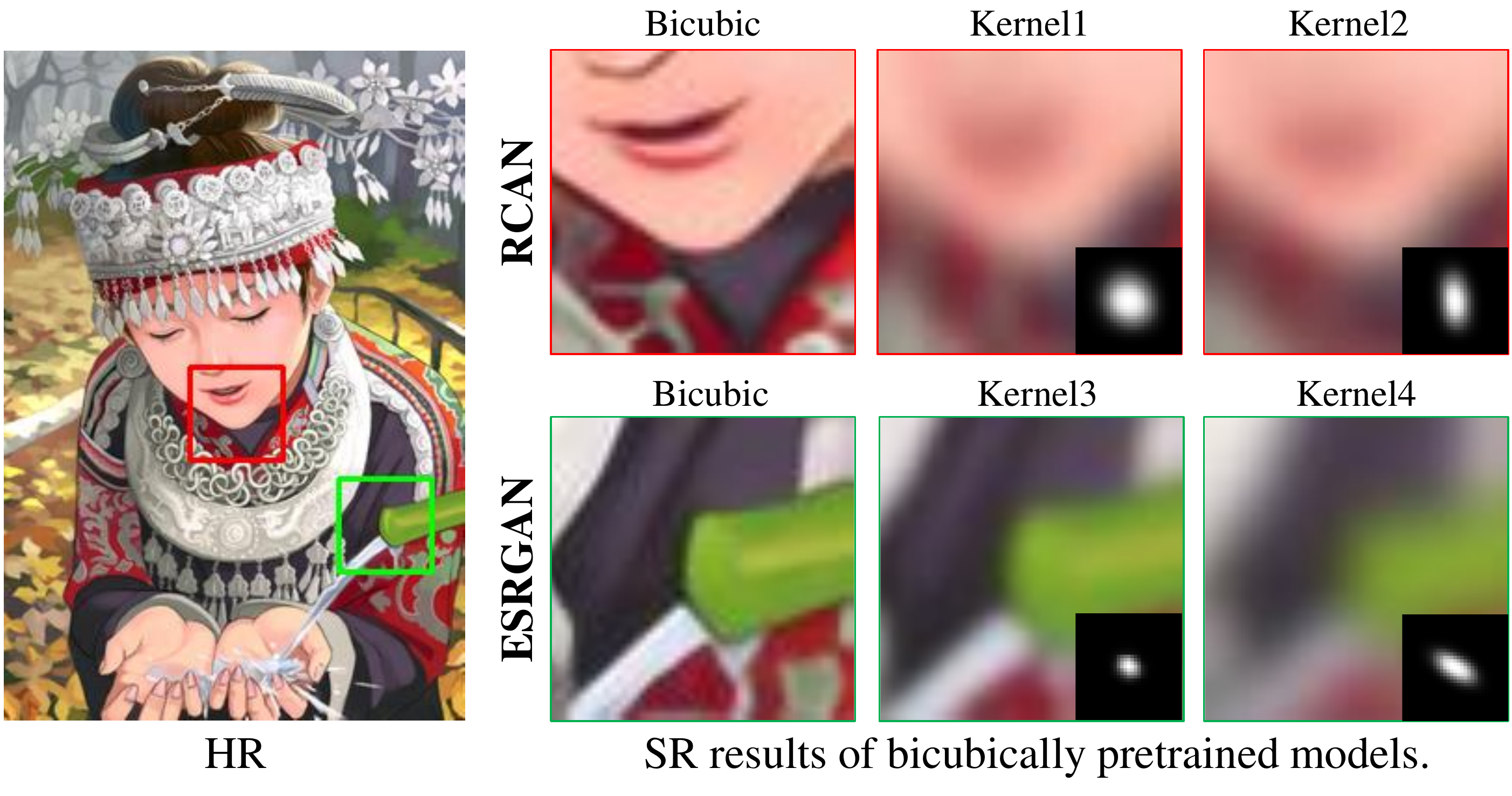}
	\caption{The adaptation problem of the offline trained ESRGAN and RCAN. The corresponding blur kernel is at the bottom right of each image.}
	\label{f:adaptation}
\end{figure}

Recently, super-resolution (SR) has been continuously advanced by various deep learning-based methods. Although these methods have exhibited promising performance, there is a common limitation: they are too 'general' and not image-specific. Firstly, they are exhaustively trained via LR-HR image pairs synthesized by predefined degradations, ignoring the real degradations of test images (\ie non-blind SR). {For example, most no-blind SR methods are optimized by paired samples synthesized with Bicubic degradation~~\cite{rcan,esrgan,casg}. And many blind SR methods also use Gaussian blur kernels to synthesize data to optimize their SR modules, such as IKC~\cite{ikc} and DASR~\cite{dasr}.} When the degradations of test images are different from the predefined ones, they may suffer a significant performance drop. Secondly, their model weights are fixed during testing, and all images are super-resolved by the same set of weights. However, test images are usually degraded by a wide range of degradations. If the model performs well for certain degradations, it is likely to perform badly for others. Thus, super-resolving all images with the same model weights may lead to sub-optimal results.
For example, as shown in Figure~\ref{f:adaptation}, ESRGAN~\cite{esrgan}, and RCAN~\cite{rcan} are trained via bicubically synthesized LR-HR pairs. They have excellent performance on bicubically downscaled images but may produce undesirable results when the images are blurred by unseen kernels. 

{Towards these issues, we propose an online updating network namely ONSR, which 1) synthesizes the training data according to the degradation of the test image, instead of by predefined degradations, and 2) updates the SR model weights for each test image instead of keeping the model fixed.  Specifically, we design two branches, namely internal branch (IB) and external branch (EB). For each given test image, IB tries to learn the internal degradation via adversarial training~\cite{gan}. With the learned degradation, EB degrades HR images from external datasets to LR-HR pairs, which are then used to update the SR module. Since the degradation is adaptively learned, the domain gap between synthesized training data and the test image could be bridged. And as the SR module can be updated for each test image, it may also get more robust to test images with a large degradation variety.}

In summary, our main contributions are as follows: 
\begin{itemize}
	\item Towards the unknown and various degradations in blind SR, we propose an online SR method. It could customize a specific model for each test LR image and thus could have more robust performance in different cases.
	
	\item We design two learning branches, IB and EB. They could learn the degradation of the given test image and adaptively update the SR module according to learned degradation. 
	
	\item Extensive experiments on both synthesized and real-world images show that ONSR can generate more visually favorable SR results and achieve state-of-the-art performance on blind SR.
\end{itemize}

\section{Related Works}
{As shown in Table~\ref{related_works}, nowadays SR  methods can be roughly divided into two categories: non-blind and blind. Blind SR methods assume that the degradation of the test image is predefined (such as bicubic downsampling) or has been estimated by degradation-prediction methods. While blind SR methods do not need extra information about the degradation. }

\begin{table}[h]
	\centering
	\caption{Examples of non-blind and blind SR methods. Non-blind SR methods assume that the degradation of  the test image is predefined or has been estimated by other methods. While blind SR methods assume that the degradation is unavailable.} \label{related_works}
	\setlength{\tabcolsep}{0.2cm}
	\resizebox{\linewidth}{!}{
		\begin{tabular}{lll}
			\toprule
			Non-blind SR methods
			&~~~~&
			Blind SR methods          \\
			\cline{1-1} \cline{3-3}
			{\begin{tabular}[l]{@{}l@{}}
					SRCNN~\cite{srcnn}, RCAN~\cite{rcan}, RRDB~\cite{esrgan}\\
					ZSSR~\cite{zssr}, SRMD~\cite{srmd}, USRNet~\cite{usr}
			\end{tabular}}
			&~~~~&
			{\begin{tabular}[l]{@{}l@{}}
					Ji \etal ~\cite{Ji2020RealWorldSV}, Cornillere \etal ~\cite{cornillere2019blind}, dSRVAE~\cite{dsrvae} \\
					KernelGAN+ZSSR~\cite{kernel_gan}, IKC~\cite{usr}, DASR~\cite{dasr}
			\end{tabular}} \\
			\bottomrule
		\end{tabular}
	}
\end{table}

\subsection{Non-Blind Super-Resolution}
In non-blind SR the degradation is predefined or known beforehand, which is easier to be studied. Thus most early SR methods are non-blind. These methods are usually trained with paired LR-HR samples synthesized by predefined degradation~\cite{simusr}, such as bicubic downsampling. Since Dong~\etal propose the first convolution neural network for SR (SRCNN) of bicubically downscaled images and achieves remarkable performance. In the following years, researchers focus on developing various network architectures. {For example, the residual dense network (RDN)~\cite{rdn} proposed by Zhang \etal and the residual-in-residual dense block (RRDB)~\cite{esrgan} proposed by Wang~\etal apply skip connections~\cite{resnet} and dense connections~\cite{densenet} to make the SR network deeper and hierarchical features more discriminative. The residual channel attention network (RCAN)~\cite{rcan} proposed by Zhang \etal, the channel attention and spatial graph convolutional network (CASG)~\cite{casg} proposed by Yang \etal, and the multi-attention augmented network (MAAN)~\cite{maan} proposed by Chen \etal all use attention mechanism to further enhance the representation capability.} Although these methods have largely advanced the SR performance for bicubic downsampling, they usually perform poorly for other degradations.  To alleviate this problem, Zhang \etal propose two networks namely SRMD~\cite{srmd} and USRNet~\cite{usr}. They input the test image and its degradation simultaneously into the network, in which way, the SR network can handle test images with various degradations. However, these methods require extra accurate degradation-estimation methods which are also left to be studied. Thus, in this paper, the proposed method focuses on blind SR, which does not need degradation prediction and is more applicable.

\subsection{Blind Super-Resolution}
Blind SR is much more challenging, as it is difficult for a single model to generalize to different degradations. In~\cite{wang2020blind}, the final results are ensembled from models that are capable of handling different cases. The SR modules in IKC~\cite{ikc}, DASR~\cite{dasr} and that proposed by Cornillere~\cite{cornillere2019blind} are all pretrained by synthesized data pairs containing a variety of degradations to be more robust to different degradations. But there are countless degradations, and we cannot train a model for each of them. Other methods try to utilize the internal prior of the test image itself. In~\cite{dpn}, the model is finetuned via similar pairs searched from the test image. Irani \etal propose ~\cite{nonpara} and KernelGAN~\cite{kernel_gan} where the blur kernel is firstly estimated by maximizing the similarity of recurring patches across multi-scale of the LR image, and then used to synthesize LR-HR training samples. Ji \etal~\cite{Ji2020RealWorldSV} also use a similar idea as KernelGAN to estimate degradations. However, the number of internal patches is limited, which heavily restricts the performance of these methods. Different from these completely internal learning-based methods, our ONSR can apply the degradation information via internal learning to external HR data, which helps to better optimize the SR module. In this way, ONSR can simultaneously take the benefits of internal and external priors in the LR and HR images respectively.

\subsection{Offline \& Online Training in Super-Resolution}
Most deep learning-based SR methods are offline optimized, \ie their model weights are only updated during training via the LR-HR pairs synthesized from external data, while keep fixed during testing. Thus, the learned model weights are completely determined by external data, without considering the inherent information of the test image. Consequently, LR images degraded by various degradations may get super-resolved by the same set of model weights. It is likely that the model only performs well on certain types of images while failing on others. Contrary to offline training, online training can get the test LR image involved in model optimization. For example, ZSSR~\cite{zssr} is an online trained SR method. It is optimized by the test LR image and its downscaled version. Therefore, it can customize the network weights for each test LR image, and could have more robust performance over different images. However, the training samples of most online trained models are limited to only one test image~\cite{nonpara,zssr, kernel_gan}. It will heavily restrict their performance. Instead, in addition to the test LR image, our ONSR can also utilize the external HR images during the online training phase. And in this way, it could better incorporate general priors of the external data and the inherent information of the test LR image.

\begin{figure}[t]
	\centering
	\includegraphics[width=\linewidth]{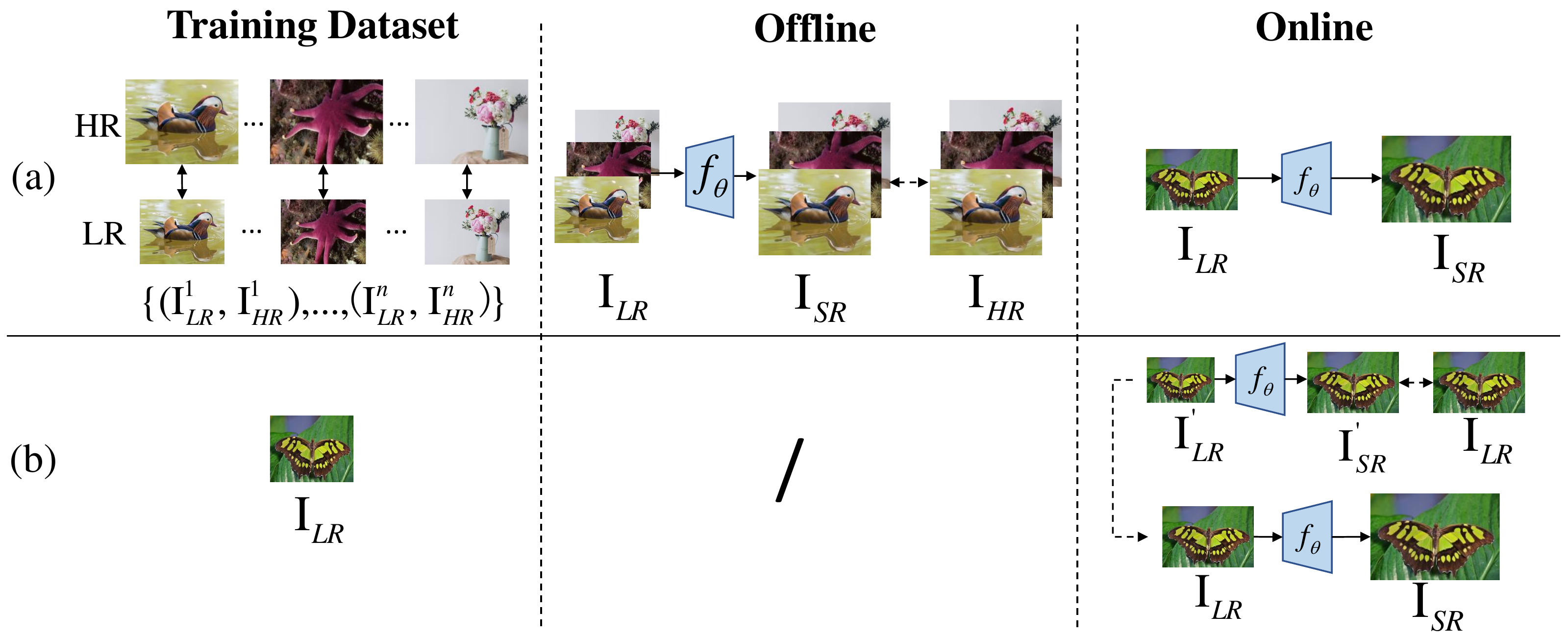}
	\caption{(a) Offline training scheme. Training datasets are synthesized from external HR image. The SR model are trained offline and only perform inference online. (b) The online training scheme of ZSSR~\cite{zssr}. Only the test image is used as the training data. The SR model is trained online.}
	\label{f:online}
\end{figure}

\section{Method}
\subsection{Motivation}\label{sec:3.1}
As we have discussed above, previous non-blind SR methods are usually offline trained (as shown in Figure~\ref{f:online}(a))~\cite{simusr}, which means LR images with various degradations are super-resolved with the same set of weights, regardless of the specific degradation of the test image. Towards this problem, a straightforward idea is to adopt an online training algorithm, \ie adjust the model weights for each test LR image with different degradations. A similar idea namely ``zero-shot" learning is used in ZSSR. As shown in Figure~\ref{f:online}(b), ZSSR is trained with the test LR image and its downscaled version. However, this pipeline has two in-born drawbacks: 1)  with a limited number of training samples, it only allows relatively simple network architectures to avoid overfitting, thus adversely affecting the representation capability of deep learning. 2) No HR images are involved. It is difficult for the model to learn general priors of HR images, which is also essential for SR reconstruction \cite{Ulyanov2018DeepIP}.

The drawbacks of ZSSR motivate us to think: a better online updating algorithm should be able to utilize both the test LR image and external HR images. The former provides inherent information about the degradation method, and the latter enables the model to exploit better general priors. Therefore, a ``general" SR model can be adjusted to process the test LR image according to its ``specific" degradation, which we call: from ``general" to ``specific".

\begin{figure}[t]
	\centering
	\includegraphics[width=\linewidth]{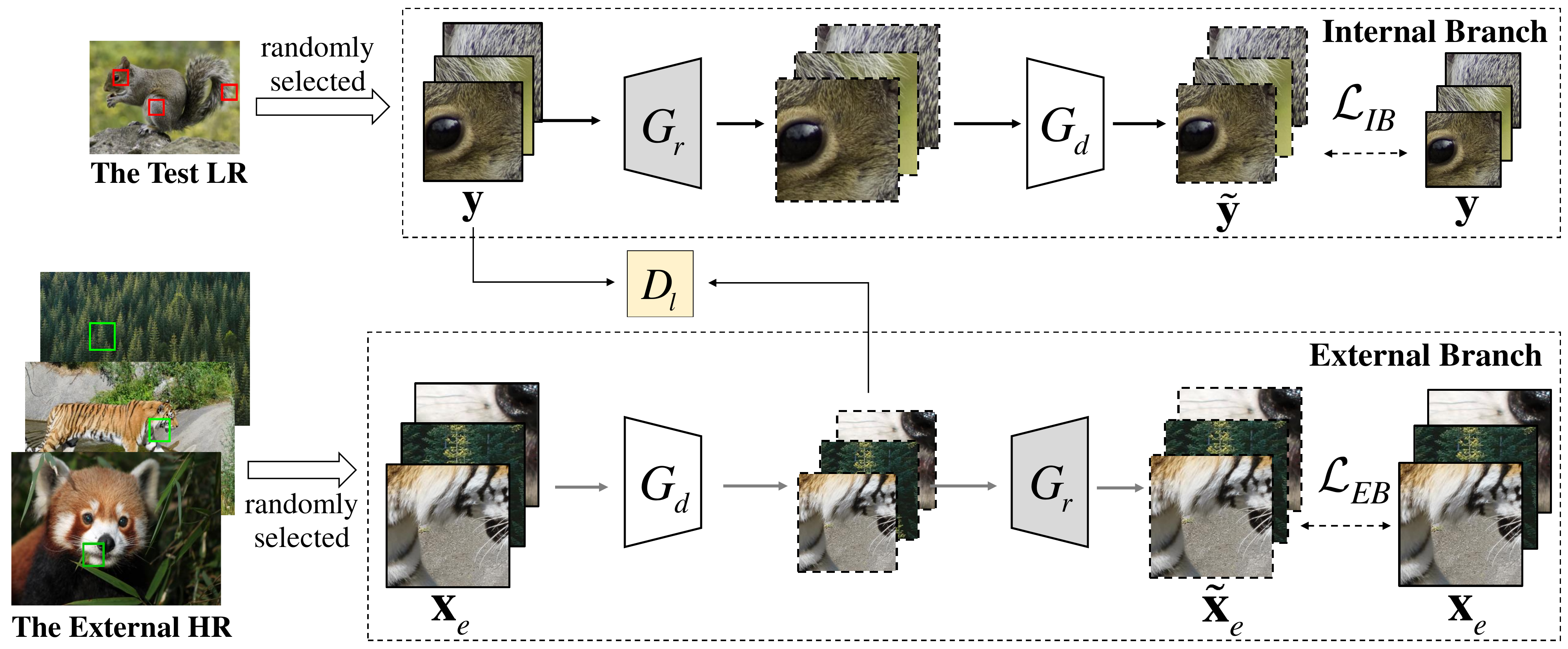}
	\caption{The online updating scheme of ONSR. Top: internal branch. Bottom: external branch. Images with solid borders are the input. Images with dotted borders are the output of $G_r$ or $G_d$.}
	\vspace{-0cm}
	\label{f:ONSR}
\end{figure}

\subsection{Formulation} \label{sec:3.2}

Accoring to the framework of MAP (maximum a posterior) \cite{ren2020neural}, the blind SR can be formulated as:
\begin{equation}\label{eq:3}
	(\mathbf{k, x}) = \mathop {\arg\max}_{\mathbf{k},\mathbf{x}} \| \mathbf{y} - (\mathbf{k}\otimes \mathbf{x})\downarrow_{s}\|^2 + \mu \phi(\mathbf{x}) + \nu \varphi(\mathbf{k})  
\end{equation}
where $\| \mathbf{y} - (\mathbf{k}\otimes \mathbf{x})\downarrow_{s}\|^2$ is the fidelity term. $\phi(\mathbf{x})$ and $\varphi(\mathbf{k})$ model the priors of sharp image and blur kernel. $\mu$ and $\nu$ are trade-off regularization parameters. Although many delicate handcrafted priors, such as the sparsity of the dark channel~\cite{darkchannel}, $L_0$-regularized intensity~\cite{LO}, and the recurrence of the internal patch~\cite{recurrence}, have been suggested for $\phi(\mathbf{x})$ and $\varphi(\mathbf{k})$, these heuristic priors could not cover more concrete and essential characteristics of different LR images. To circumvent this issue, we design two modules, \ie the reconstruction module $G_r$ and the degradation estimation module $G_d$, which can capture priors of $\mathbf{x}$ and $\mathbf{k}$ in a learnable manner.  We substitute $\mathbf{x}$ by $G_r(\mathbf{y})$, and denote the degradation process as $G_d(\cdot)$, then the problem becomes:
\begin{equation}
	\mathop{\arg\min}_{\theta_{G_r},\theta_{G_d}}\| \mathbf{y} - G_d(G_r(\mathbf{y}; \theta_{G_r}); \theta_{G_d}) \|,
\end{equation}
The prior terms are removed because they could also be captured by the generative networks $G_r(\cdot)$ and $G_d(\cdot)$~\cite{Ulyanov2018DeepIP}. 

This problem involves the optimization of two neural networks, \ie $G_r$ and $G_d$. Thus, we can adopt an alternating optimization strategy:
\begin{equation}
	\left \{
	\begin{aligned}
		&\theta_{G_r}^{i+1}= \mathop{\arg\min}_{\theta_{G_r}}\| \mathbf{y} - G_d (G_r(\mathbf{y}; \theta_{G_r}); \theta_{G_d}^i ) \|\\
		&\theta_{G_d}^{i+1}=\mathop{\arg\min}_{\theta_{G_d}}\| \mathbf{y} - G_d (G_r(\mathbf{y}; \theta_{G_r}^i); \theta_{G_d}) \|.\\
	\end{aligned}
	\right.
\end{equation}
In the first step, we fix $G_d$ and optimize $G_r$, while in the second step we fix $G_r$ and optimize $G_d$. 

So far only the given LR image is involved in this optimization. However, as we have discussed in Sec~\ref{sec:3.1}, the limited training sample may be not enough to get $G_r$ sufficiently optimized, because there are usually too many learnable parameters in $G_r$. Thus, we introduce the external HR images $\mathbf{x}_e$ in the optimization of $G_r$. In the $i^{th}$ step, we degrade the $\mathbf{x}_e$ by $G_d(\ \cdot\  ; \theta_{G_d}^i)$ to $\mathbf{y}_e$. Then $\mathbf{x}_e$ and $\mathbf{y}_e$ could form a paired sample that could be used to optimize $G_r$. Thus, the alternating optimization process becomes:
\begin{equation}
	\left \{
	\begin{aligned}
		&{\mathbf{y}_e} = G_d({\mathbf{x}_e}; \theta_{G_d}^i) \\
		&\theta_{G_r}^{i+1}= \mathop{\arg\min}_{\theta_{G_r}}\| {\mathbf{x}_e} - G_r({\mathbf{y}_e};\theta_{G_r})\|\\
		&\theta_{G_d}^{i+1}=\mathop{\arg\min}_{\theta_{G_d}}\| \mathbf{y} - G_d (G_r(\mathbf{y}; \theta_{G_r}^i); \theta_{G_d}) \|, \\
	\end{aligned}
	\right.
\end{equation}
in which, $G_r$ is optimized by external datasets, while $G_d$ is optimized by the given LR image only. At this point, we have derived the proposed method from the perspective of alternating optimization. This may help better understand OSNR.

\subsection{Online Super-Resolution} \label{sec:3.3}
As illustrated in Figure~\ref{f:ONSR},  our online SR (ONSR) consists of two branches, \ie IB and EB.  IB and EB share the  reconstruction module $G_r$ and the degradation estimation module $G_d$. $G_r$ aims to map the given LR image $\mathbf{y}$ from the LR domain $\mathcal{Y}\subset \mathbb{R}^{3\times H \times W}$ to the HR domain $\mathcal{X} \subset \mathbb{R}^{3\times sH \times sW}$, \ie reconstructing an SR image $\mathbf{x}$. While $G_d$ aims to estimate the specific degradation of the test LR image.  

In IB, only the given LR image is involved. As shown in Figure~\ref{f:ONSR}, the input of IB are patches randomly selected from the test LR image. The input LR patch $\mathbf{y} \sim p_{\mathcal{Y}}$ is firstly super resolved by $G_r$ to an SR patch. Then this SR patch is further degraded by $G_d$ to a fake LR patch. To guarantee that the fake LR can be translated to the original LR domain, It is supervised by the original LR patch via L1 loss. The paired SR and LR patches could help $G_d$ to learn the specific degradation of the test image. The optimization details will be further explained in Section~\ref{sec:3.4}.                   

In EB, only external HR images are involved. The input of EB are patches randomly selected from different external HR images. Conversely, the external patch $\mathbf{x}_e\sim p_{\mathcal{X}}$ is firstly degraded by $G_d$ to a fake LR patch, . As the weights of  $G_d$ are shared between IB and EB, the external patches are actually degraded by the learned degradation. Thus, the paired HR and fake LR patches could help $G_r$ learn to super resolve LR images with specific degradations.

According to the above analysis, the loss functions of IB and EB can be formulated as:
\begin{equation}\label{eq:4}
	\mathcal{L}_{IB} = \mathbb{E}_{\mathbf{y}\sim p_{\mathcal{Y}}}\| \mathbf{y} - G_d(G_r(\mathbf{y}; \theta_{G_r}); \theta_{G_d}) \|_1,
\end{equation} 
\begin{equation}\label{eq:5}
	\mathcal{L}_{EB} = \mathbb{E}_{\mathbf{x}_e\sim p_{\mathcal{X}}}\| \mathbf{x}_e - G_r(G_d(\mathbf{x}_e; \theta_{G_d}); \theta_{G_r}) \|_1.
\end{equation} 	

Since information in the single test LR image is limited, to help $G_d$ better learn the specific degradation, we further adopt the adversarial learning strategy. As shown in Figure~\ref{f:ONSR}, we introduce a discriminator $D_l$. $D_l$ is used to discriminate the distribution characteristics of the LR image. It could force $G_d$ to generate fake LR patches that are more similar to the real ones. Thus more accurate degradations could be learned by $G_d$. We use the GAN formulation as follows,
\begin{equation}\label{eq:6}
	\begin{aligned}
		\mathcal{L}_{GAN} = \mathbb{E}_{\mathbf{y}\sim p_\mathcal{Y}}[logD_l(\mathbf{y}; \theta_{D_l})] + \mathbb{E}_{\mathbf{x}_e\sim p_{X}}[log(1-D_l(G_d(\mathbf{x}_e; \theta_{G_d}); \theta_{D_l}))].
	\end{aligned}
\end{equation}
Adversarial training is not used for the intermediate output $G_r(\mathbf{y})$, because it may lead $G_r(\mathbf{y})$ to generate unrealistic textures~\cite{esrgan}.We also experimentally explain this problem in Section~\ref{sec:4.4.3}.

\subsection{Separate Optimization} \label{sec:3.4}
Generally, most SR networks are optimized by the weighted sum of all objectives. All modules in an SR network are treated indiscriminately. Unlike this commonly used joint optimization method, we propose a separate optimization strategy. Specifically, 
$G_d$ is optimized by the objectives that are directly related to the test LR image, while $G_r$ is optimized by objectives that are related to external HR images. The losses for these two modules are as follows,
\begin{equation}\label{eq:7}
	\mathcal{L}_{G_d} = \mathcal{L}_{IB}+\lambda\mathcal{L}_{GAN}
\end{equation}
\begin{equation}\label{eq:8}
	\mathcal{L}_{G_r} = \mathcal{L}_{EB}
\end{equation}
where $\lambda$ controls the relative importance of the two losses.
\begin{algorithm}[H] 
	\caption{Algorithm of ONSR}
	\label{alg:1}
	\begin{algorithmic}[1] 
		\Require 
		{The LR image to be reconstructed: $I_{LR}$} 
		\Statex \hspace{0.4cm} \ The external HR image dataset: $\boldsymbol{S}_{HR}$
		\Statex \hspace{0.4cm} \ Maximum updating step: $T$
		\Statex \hspace{0.4cm} \ Online testing step interval: $t$
		\Ensure
		The best SR image: $\boldsymbol{I}_{SR}=G_r(I_{LR};\ g_r )$
		\State Load the pretrained model $G_r$
		\State $i=0$ 
		
		\While {$i\leq T$} 
		\State {$i \leftarrow{i+1}$}
		\State Sample $n$ LR image patches ${{\mathbf{\{y}^j\}}}_{j=1}^n$ from  $I_{LR}$ 
		\State Sample $n$ HR image patches ${{\mathbf{\{x}^j_e\}}}_{j=1}^n$ from $\boldsymbol{S}_{HR}$
		\State // \textit{Online testing}
		\If {$i\%t==0$}
		\State $I_{SR}^i = G_r^i(I_{LR};\ g_r^i)$ 
		\EndIf
		\State // \textit{Online updating different modules}
		
		\State Update $\theta_{G_d} \leftarrow \theta_{G_d} - (\nabla_{\theta_{G_d}} \mathcal{L}_{IB} + \lambda \nabla_{\theta_{G_d}}\mathcal{L}_{GAN})$
		\vspace{0.05cm}	   	
		\State Update $\theta_{G_r} \leftarrow \theta_{G_r} -  \nabla_{\theta_{G_r}}\mathcal{L}_{EB} $
		\vspace{0.05cm}
		\State Update $\Delta \theta_{D_l} \leftarrow \theta_{D_l} -  \lambda\nabla_{\theta_{D_l}}\mathcal{L}_{GAN}$
		
		\EndWhile 
		\label{code:recentEnd} 
	\end{algorithmic} 
\end{algorithm}
We adopt this separate optimization strategy for two reasons. Firstly, as the analysis in Section~\ref{sec:3.2} that $G_d$ and $G_r$ are alternately optimized in ONSR, separate optimization may make these modules easier to converge~\cite{usr}.  Secondly, $G_d$ aims to learn the specific degradation of the test image, while $G_r$ needs to learn the general priors from external HR images. Thus it is more targeted for them to be separately optimized. We experimentally prove the superiority of separate optimization in Sec~\ref{exp_sep_opt}. The overall algorithm is shown in Algorithm~\ref{alg:1}.

\begin{figure}[t]
	\centering
	\includegraphics[width=\linewidth]{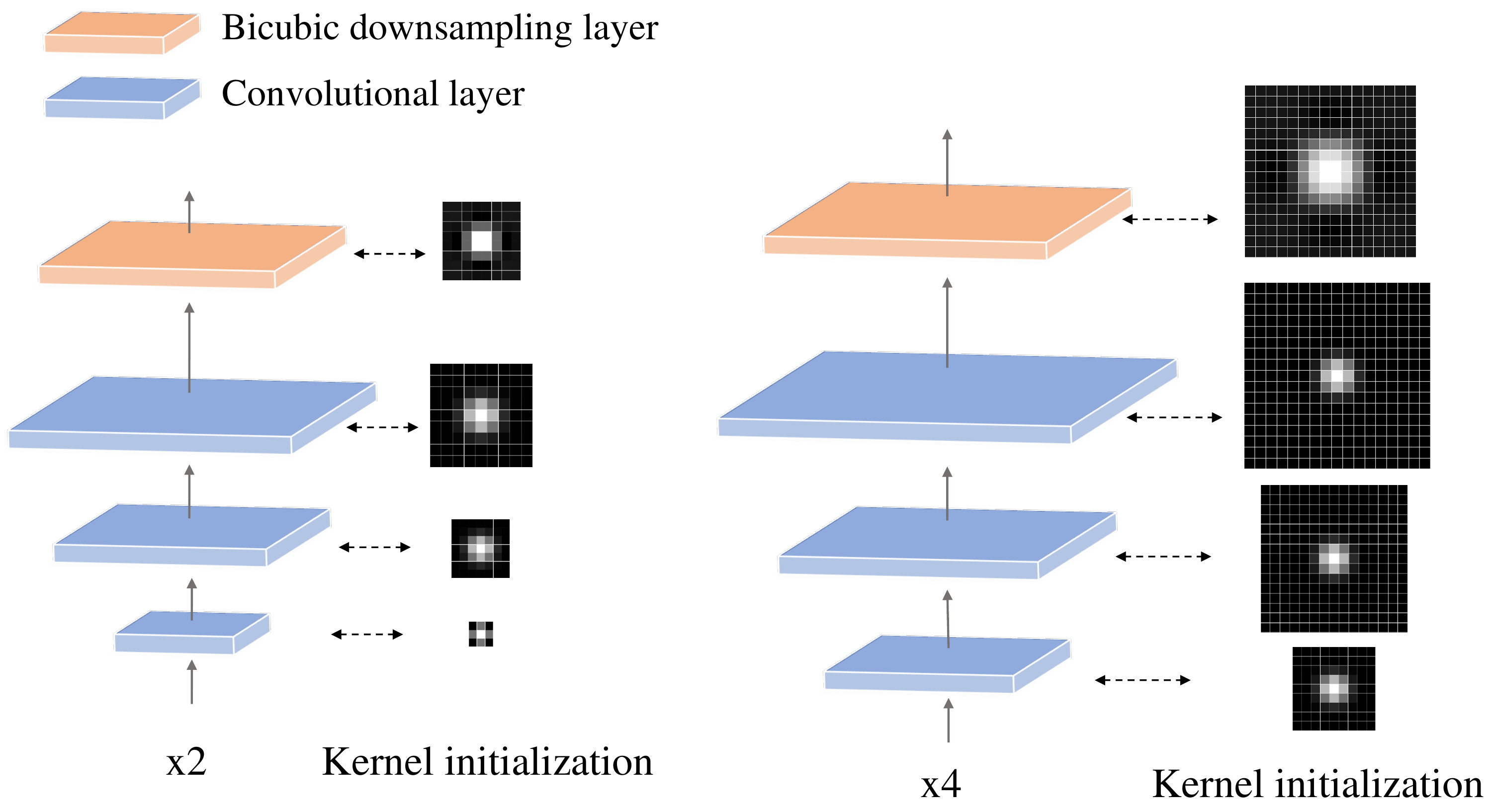}
	\caption{The architecture and initialization of $G_d$.}
	\label{f:GD}
\end{figure}

\subsection{Network Instantiation} \label{sec:3.5}
Most existing SR structures can be used as $G_r$  and integrated into ONSR. In this paper, we mainly use Residual-in-Residual Dense Block (RRDB)~\cite{esrgan}. RRDB combines the multi-level residual network and dense connections, which is easy to be trained and has promising performance on SR.  $G_r$  consists of  23 RRDBs and an upsampling module. It is initialized using the pre-trained network parameters, which could render additional priors of external data, and also provide a comparatively reasonable initial point to accelerate optimization.

The architecture of the degradation module $G_d$ is shown in Figure~\ref{f:GD}. We can see that blurring and downsampling are linear transforms in Eq.~\ref{eq:1}, so $G_d$ is designed as a deep linear network. Theoretically, a single convolutional layer should be able to represent all possible downsampling blur methods in Eq.~\ref{eq:1}. However, linear networks have infinitely many equal global minimums according to~\cite{Arora2018OnTO}, which makes the gradient-based optimization faster for deeper linear networks than shallower ones. Thus, we employ three convolutional layers with no activations and a bicubic downsampling layer in $G_d$. The bicubic downsampling layer could help obtain a reasonable initial point, which is similar to that in~\cite{kernel_gan} but simpler. Additionally, to accelerate the convergence of $G_d$, we use isotropic Gaussian kernels with a standard deviation of 1 to initialize all convolutional layers, as shown in Figure~\ref{f:GD}. Considering that images with larger downsampling factor are usually more seriously degraded, we set the size of the three convolutional layers to $3\times3$, $7\times7$, $9\times9$ for scale factor $\times2$, and $9\times9$, $15\times15$, $17\times 17$ for scale factor $\times 4$.

$D_l$ is a VGG-style network \cite{simonyan2014very} to perform discrimination. The input size of $D_l$ is $32\times32$.

\begin{figure}[H]
	\centering
	\includegraphics[width=\linewidth]{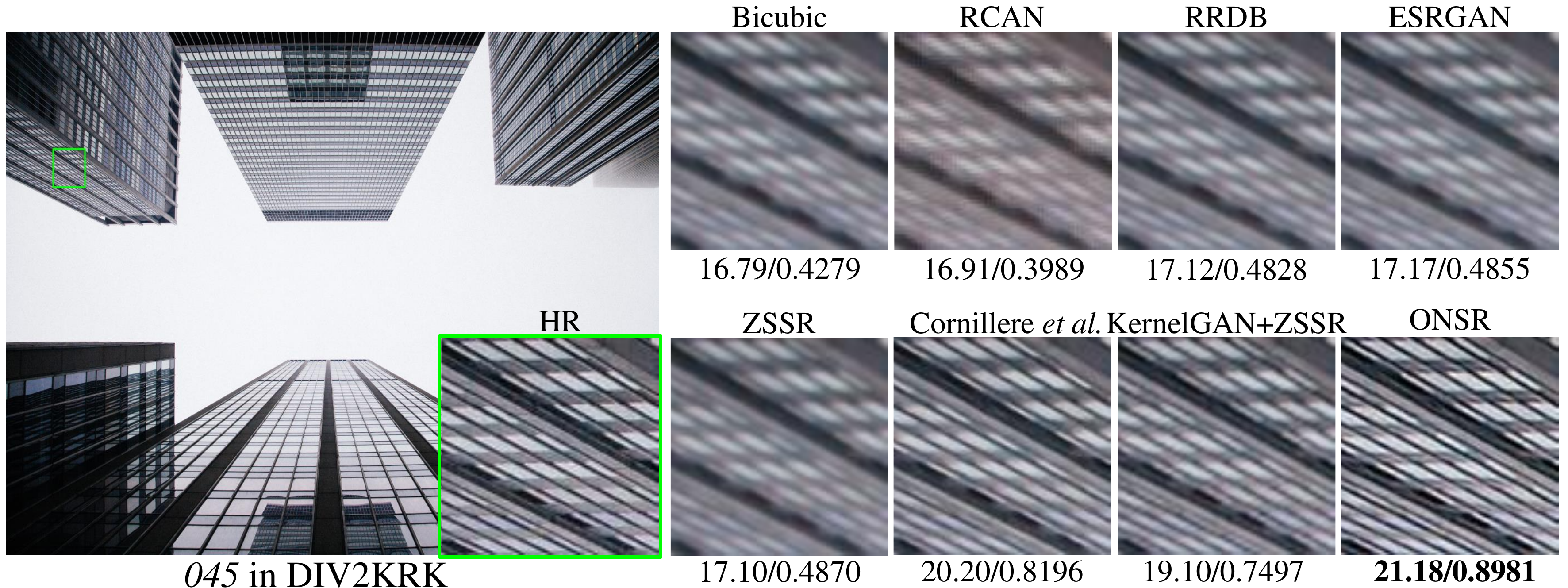}
	\includegraphics[width=\linewidth]{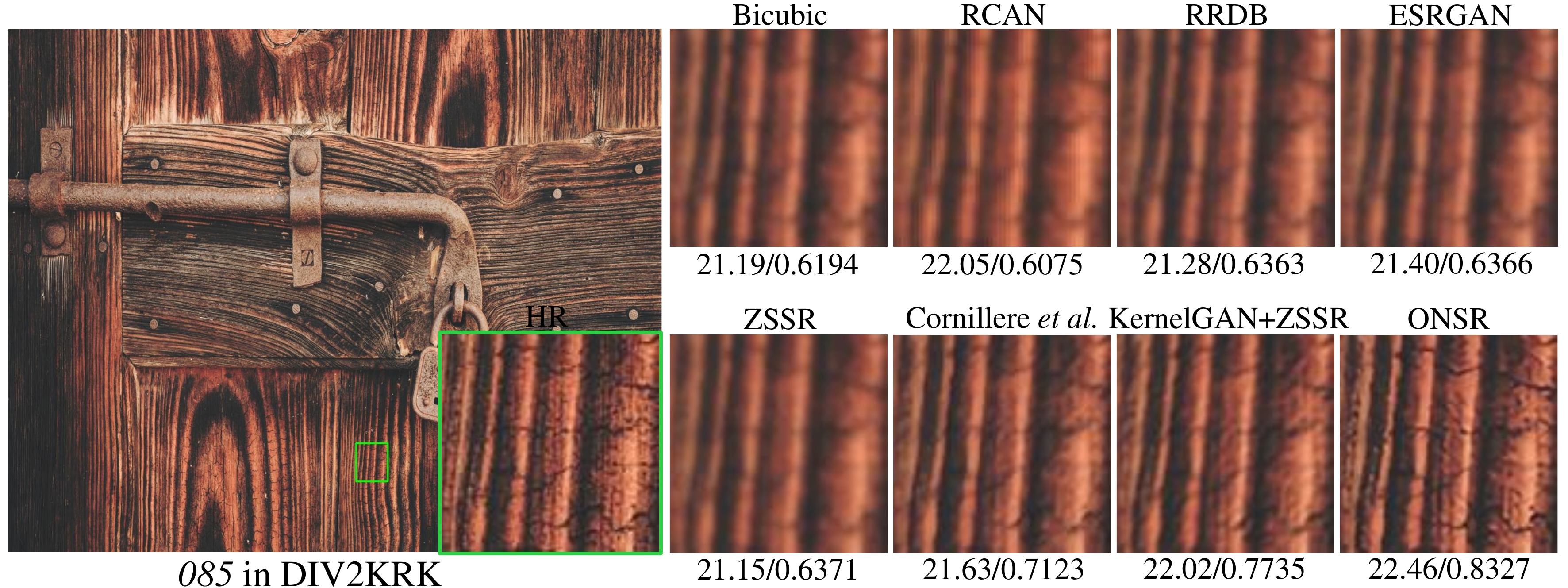}
	\includegraphics[width=\linewidth]{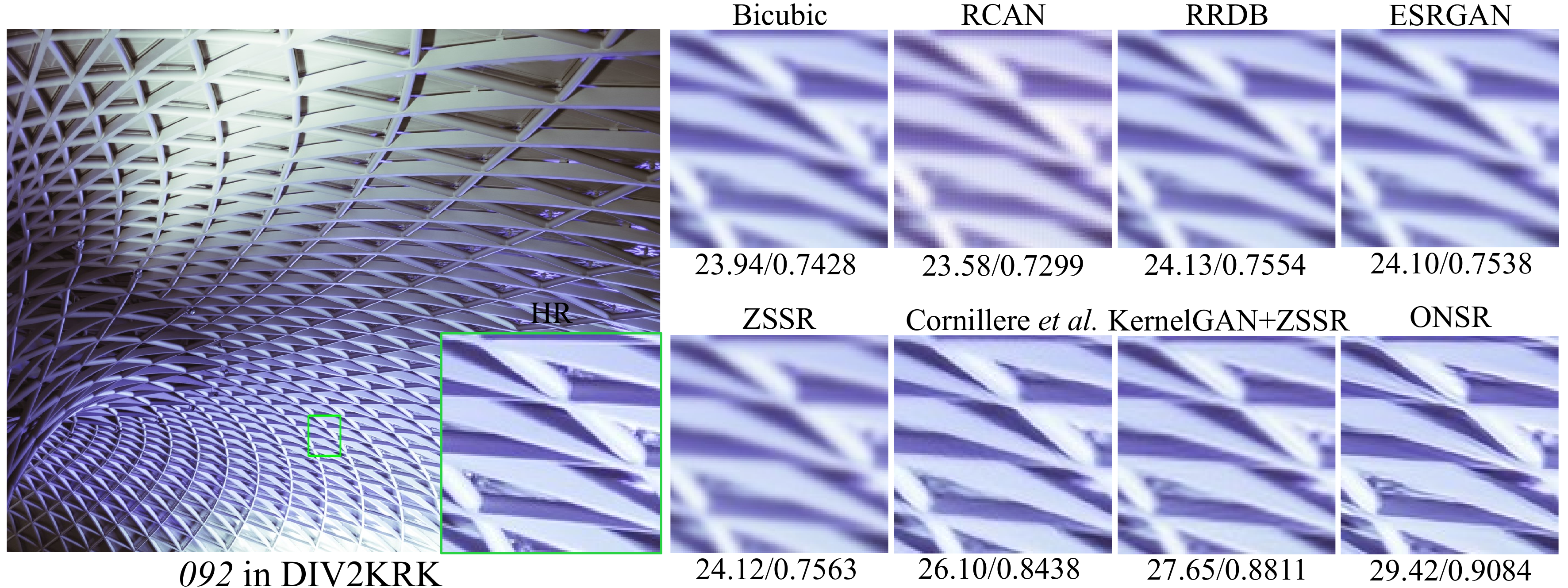}
	\caption{Visual comparison of ONSR and SotA SR methods for $2\times$ SR. The model name is denoted above the corresponding patch and PSNR/SSIM is denoted below.}
	\label{f:visual_result}
	\vspace{-0cm}
\end{figure}

\begin{figure}[H]
	\centering
	\includegraphics[width=\linewidth]{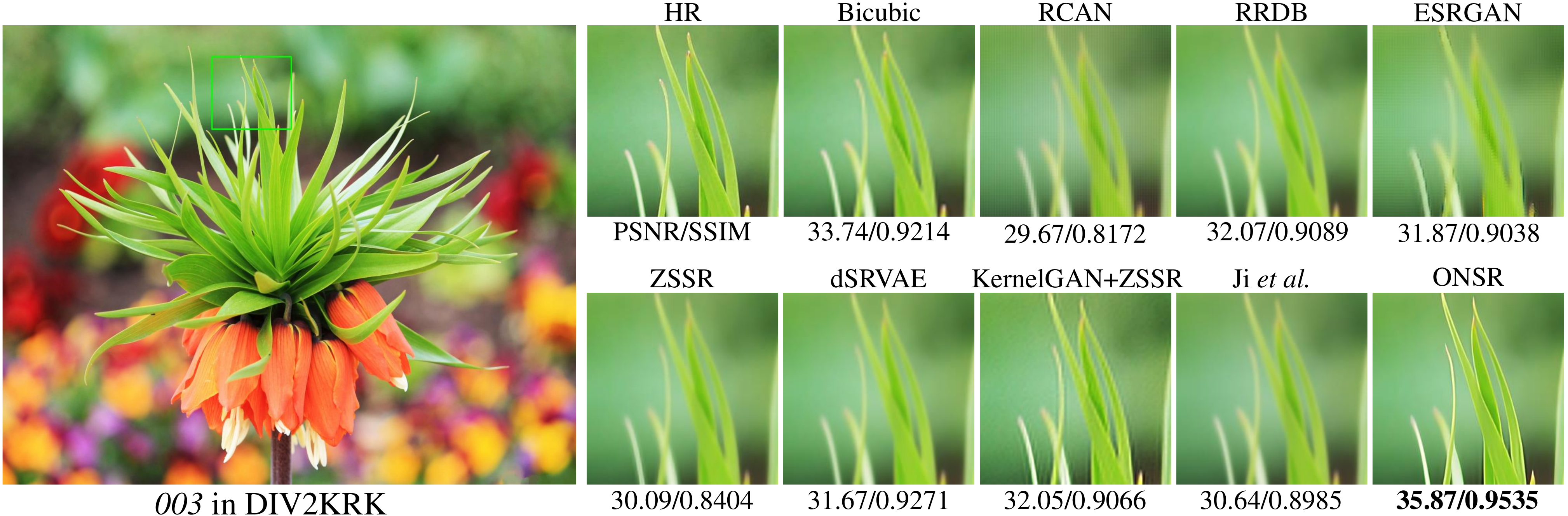}
	\includegraphics[width=\linewidth]{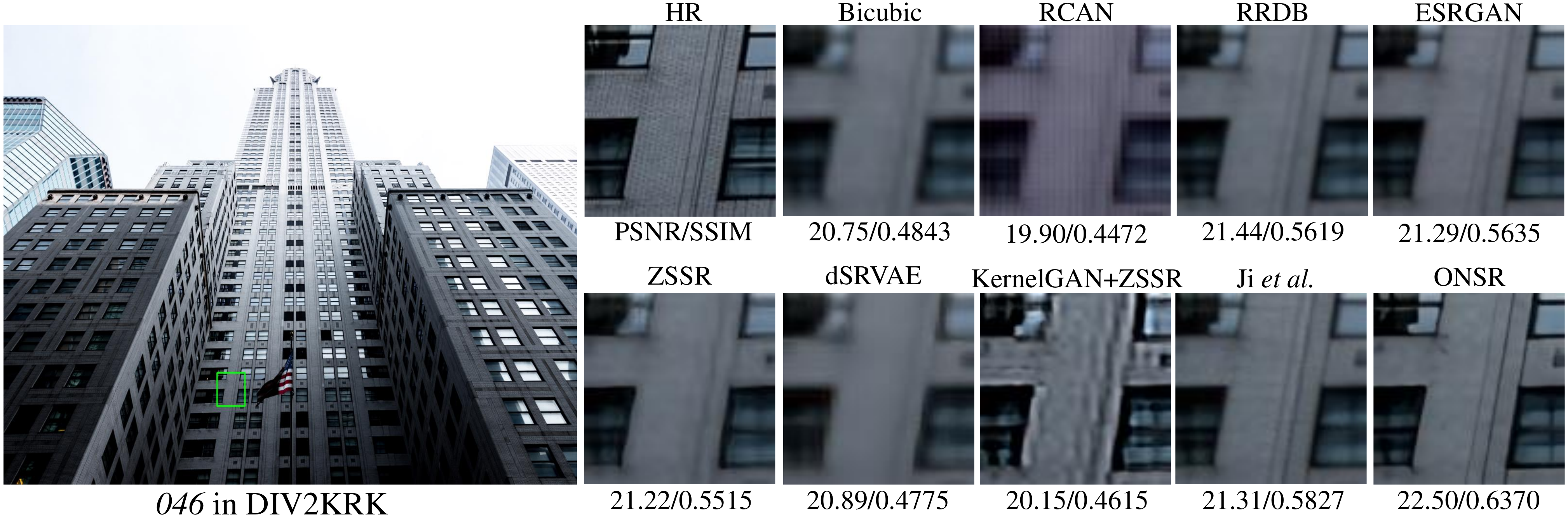}
	\includegraphics[width=\linewidth]{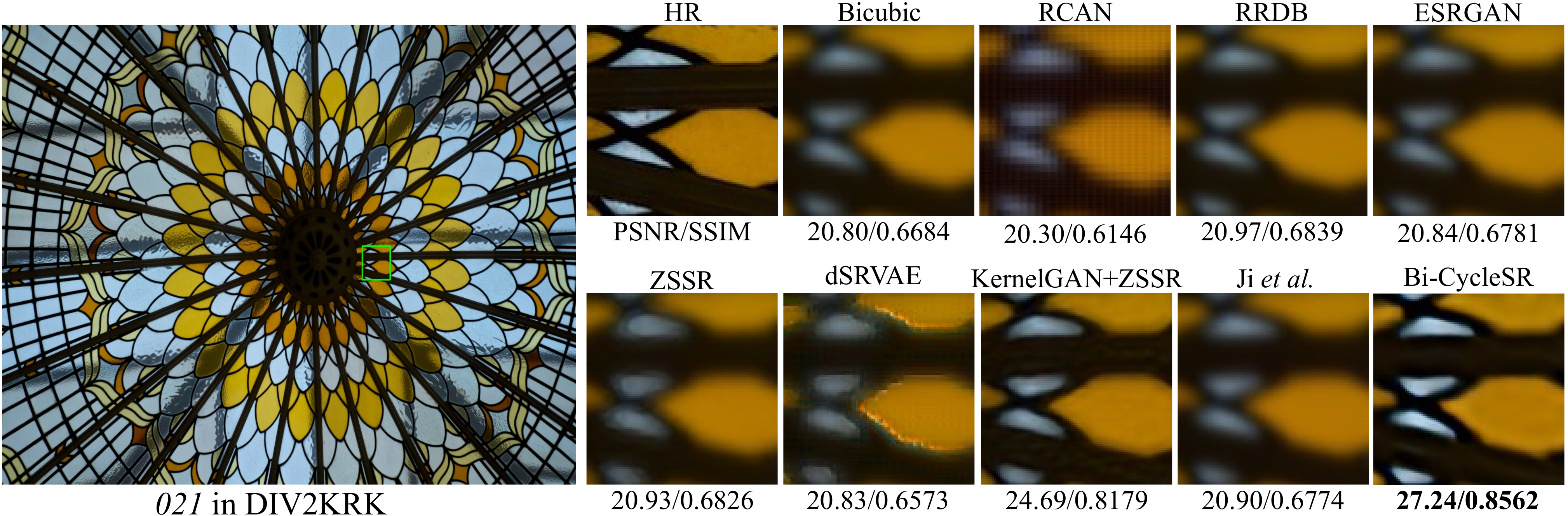}
	\caption{Visual comparison of ONSR and SotA SR methods for $4\times$ SR. The model name is denoted above the corresponding patch and PSNR/SSIM is denoted below.}
	\label{f:visual_result1}
	\vspace{-0cm}
\end{figure}

\section{Experiments}

\subsection{Experimental Setup}\label{sec:4.1}
\textbf{Datasets.} We use 800 HR images from the training set of DIV2K \cite{div2k} as the external HR dataset and evaluate the SR performance on \textit{DIV2KRK} \cite{kernel_gan}. LR images in DIV2KRK are generated by blurring and subsampling each image from the validation set (100 images) of DIV2K with randomly generated kernels. These kernels are isotropic or anisotropic Gaussian kernels with random lengths $\lambda_1, \lambda_2 \sim \mathcal{U}(0.6,5)$ independently distributed for each axis, rotated by a random angle $\theta\sim\mathcal{U}[-\pi, \pi]$. To deviate from a regular Gaussian kernel, uniform multiplicative noise (up to $25\%$ of each pixel value of the kernel) is further applied.

\textbf{Evaluation Metrics.} To quantitatively compare the SR performance different methods, we use PSNR, SSIM~\cite{wang2004image}, Perceptual Index (PI) \cite{blau2018the} and Learned Perceptual Image Patch Similarity(LPIPS) \cite{zhang2018the}. Contrary to PSNR and SSIM, lower PI and LPIPS indicate higher perceptual quality.

\textbf{Training Details.} We randomly sample 10 patches of $32\times32$  from the LR image and 10 patches of $32s\times 32s$ from different HR images for each input minibatch, where $s$ denotes the scaling factor. ADAM~\cite{adam} optimizer with $\beta_1=0.9, \beta_2=0.999$ is used for optimization.
We set the online updating step to 500 for each image, and the LR image is tested every 10 steps. To accelerate the optimization, we initialize ONSR with the bicubically pretrained model of RRDB, which is publicly available.

\begin{table}[t]
	\caption{Quantitative comparison on DIV2KRK dataset. $\times2$ and $\times4$ denote the scale factors. $\uparrow$ denotes the larger the better. $\downarrow$ denotes the smaller the better. Best and second best results are \textbf{highlighted} and \underline{underlined}.}
	\label{t:sota}
	\centering
	\setlength{\tabcolsep}{0.15cm}
	\resizebox{\linewidth}{!}{
		\begin{tabular}{llccccccccc}
			\toprule
			\multirow{2}{*}{Type} 
			& \multirow{2}{*}{Method}
			& \multicolumn{4}{c}{$\times$2}
			&
			&\multicolumn{4}{c}{$\times$4}\\
			\cline{3-6}\cline{8-11}
			&& PSNR $\uparrow$
			& SSIM $\uparrow$& PI $\downarrow$ 
			& LPIPS $\downarrow$
			&
			& PSNR $\uparrow$ 
			& SSIM $\uparrow$
			& PI $\downarrow$
			& LPIPS $\downarrow$ \\ 
			\midrule
			\multirow{5}{*}{\begin{tabular}[c]{@{}l@{}}Type 1:\\  Non-BlindSR\end{tabular}}
			& Bicubic 
			& 28.81             & 0.8090             & 6.7039             & 0.3609
			&& 25.46             & 0.6837             & 8.6414             & 0.5572             \\
			& ZSSR \cite{zssr} 
			& 29.09             & 0.8215             & 6.2707             & 0.3252
			&& 25.61             & 0.6920             & 8.1941             & 0.5192             \\
			& ESRGAN \cite{esrgan} 
			& 29.18             & 0.8212             & 6.1826             & 0.3178
			&&25.57             & 0.6906             & 8.3554             & 0.5266             \\
			& RRDB \cite{esrgan}
			& 29.19             & 0.8224             & 6.4801             & 0.3376
			&& 25.66             & 0.6937             & 8.5510             & 0.5416             \\
			& RCAN \cite{rcan}
			& 27.94             & 0.7885             & 6.8855             & 0.3417
			&& 24.75             & 0.6337             & 8.4560             & 0.5830             \\
			\midrule
			\multirow{4}{*}{\begin{tabular}[c]{@{}l@{}}Type 2:\\ BlindSR\end{tabular}}
			& Cornillere \etal \cite{cornillere2019blind}
			& 29.42             & 0.8459             & 4.8343             & \textbf{0.1957}
			&& -                   & -                    & -                    & -                    \\
			& dSRVAE \cite{dsrvae}
			& -                   & -                    & -                    & - 
			&& 25.07             & 0.6553             & \textbf{5.7329}  & 0.4664             \\
			& Ji \etal \cite{Ji2020RealWorldSV}
			& -                   & -                    & -                    & -
			&& 25.41             & 0.6890             & 8.2348             & 0.5219             \\
			& KernelGAN+ZSSR \cite{kernel_gan}
			& \underline{29.93} & \underline{0.8548} & \underline{5.2483} &0.2430
			&& \underline{26.76} & \underline{0.7302} & 7.2357             &\underline{0.4449} \\
			& ONSR (Ours)
			& \textbf{31.34}  & \textbf{0.8866}  & \textbf{4.7952}  & \underline{0.2207}
			&& \textbf{27.66}  & \textbf{0.7620}  & \underline{7.2298} & \textbf{0.4071}  \\ 
			\bottomrule
		\end{tabular}
	}
\end{table}

\subsection{Super-Resolution on Synthetic Data}
We compare ONSR with other state-of-the-art (SotA) methods on the synthetic dataset DIV2KRK. We present two types of algorithms for analysis: 1) \textit{Type}1 includes ESRGAN \cite{esrgan}, RRDB \cite{esrgan}, RCAN \cite{rcan} and ZSSR \cite{zssr}. They are non-blind SotA SR methods trained on bicubically downsampled images. 2) \textit{Type} 2 are blind SR methods including KernelGAN+ZSSR \cite{kernel_gan}, dSRVAE \cite{dsrvae}, Ji \etal \cite{Ji2020RealWorldSV} and Cornillere \etal \cite{cornillere2019blind}.

\textbf{Quantitative Results.} SotA non-bind SR methods such as ESRGAN and RRDB have remarkable performance on bicubically downscaled images. However, as shown in Table~\ref{t:sota} \textit{Type} 1, when tested on DIV2KRK, they perform only slightly better than the naive bicubic interpolation. And RCAN is even worse. { The performance drop suggests that these methods fail to generalize to the test images with various degradations in DIV2KRK}. As a comparison, our ONSR outperforms these non-blind SR methods by a large margin. Specifically, RRDB shares the same network architectures with $G_r$ in ONSR, while ONSR outperforms it by \textbf{2.15 dB} and \textbf{2 dB} for scales $\times2$ and $\times4 $ respectively.  { This comparison demonstrates the effectiveness of the online updating strategy adopted by ONSR}. 

{ As shown in Table~\ref{t:sota} \textit{Type} 2, ONSR also shows its superiority over other blind SR methods. KernelGAN+ZSSR is a blind SR method that also adopts an online updating strategy. However, as we have discussed in Sec~\ref{sec:3.1}, it does not exploit information from the external dataset, which may restrict its performance}. As a result,  ONSR outperforms it by \textbf{1.41 dB} and \textbf{0.90 dB} for scales $\times2$ and $\times4 $ respectively.  {This comparison indicates that ONSR successfully integrates information from both internal and external branches and achieves better SR performance}.

\textbf{Qualitative Results.} In Figure~\ref{f:visual_result} and \ref{f:visual_result1}, we  present visual comparisons between these methods on scales $\times 2$ and $\times 4$ respectively. { As shown in these figures,  the images produced by SotA non-blind SR methods are not sharp enough. RCAN even changes the colors of the original images. The results produced by blind SR methods are relatively sharper, but they also tend to contain undesirable artifacts and distortions, such as the window contours in image \textit{046}}. As a comparison, the SR images produced by ONSR are cleaner and more visually favorable.

\begin{figure}[H]
	\centering
	\includegraphics[width=\linewidth]{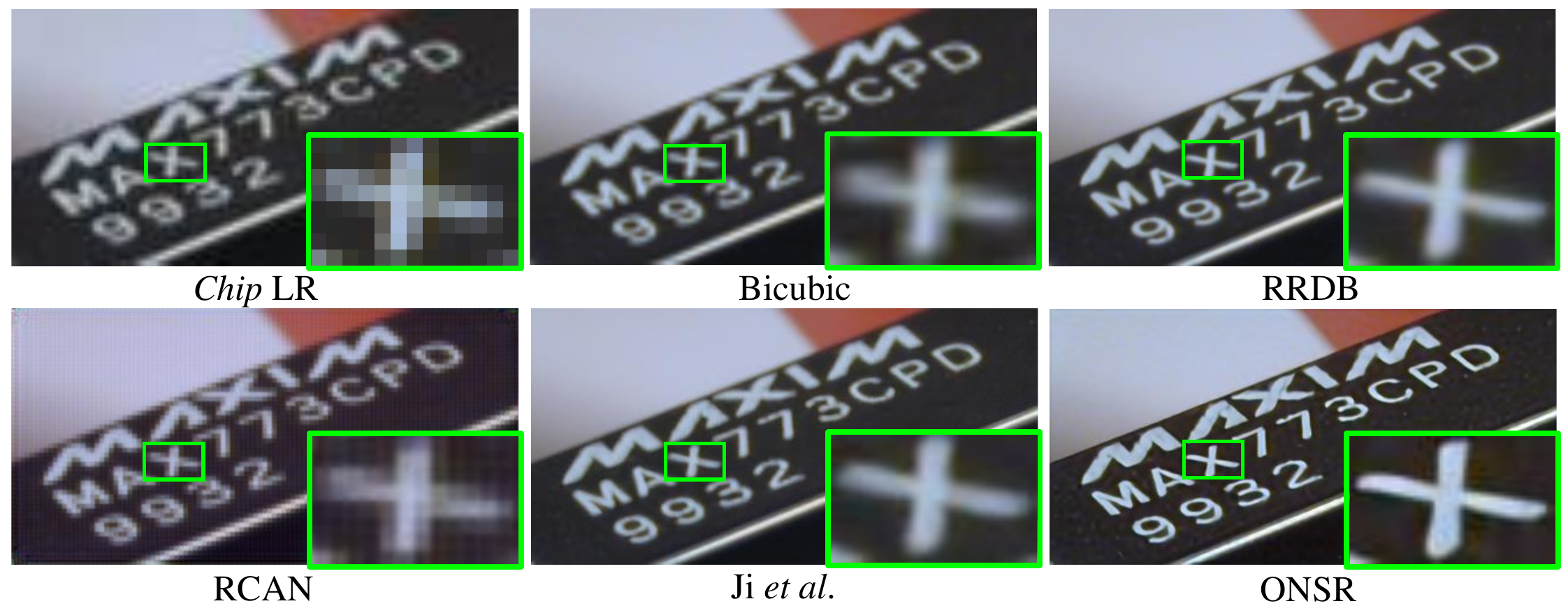}\vspace{2mm}
	\includegraphics[width=\linewidth]{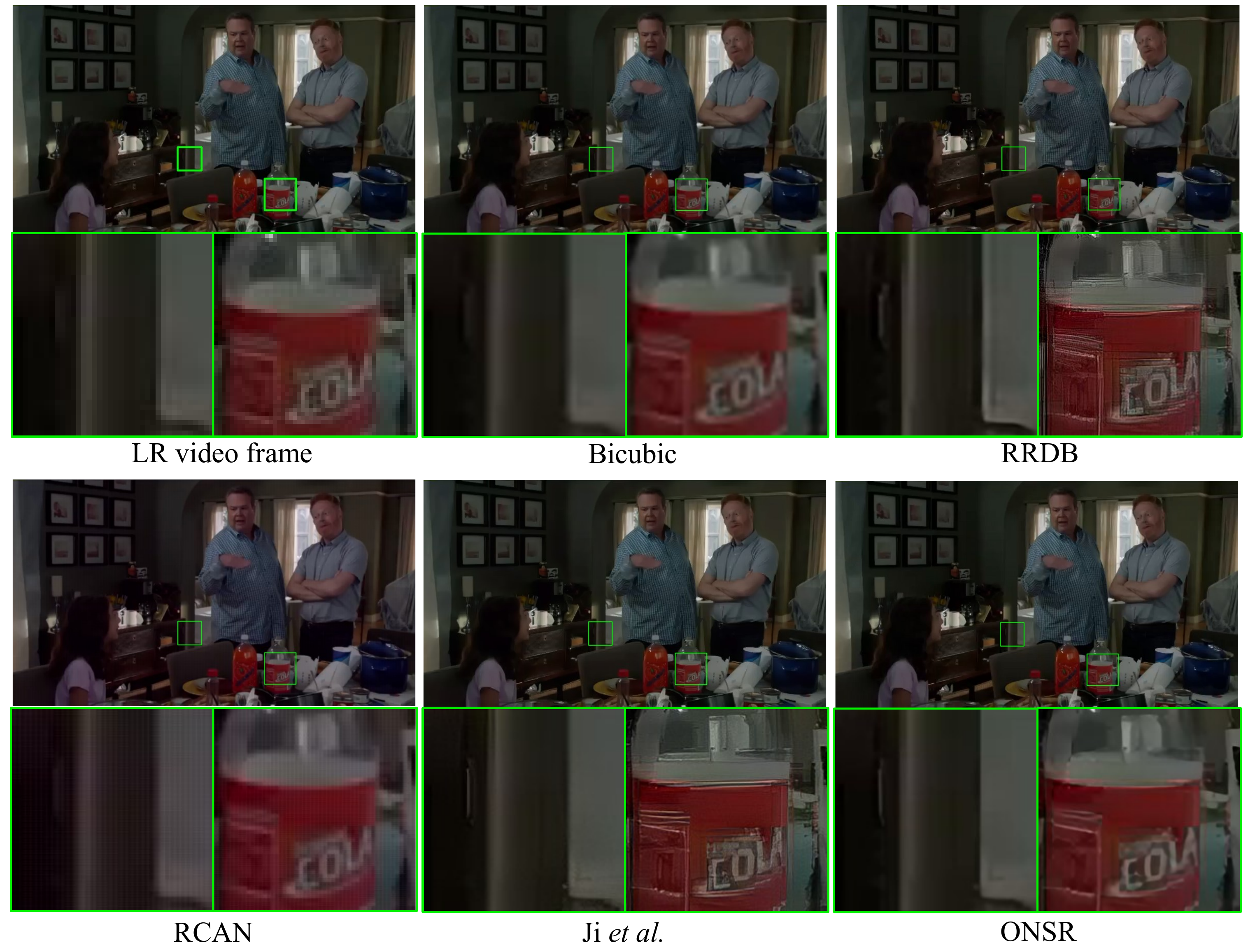}\vspace{2mm}
	\includegraphics[width=\linewidth]{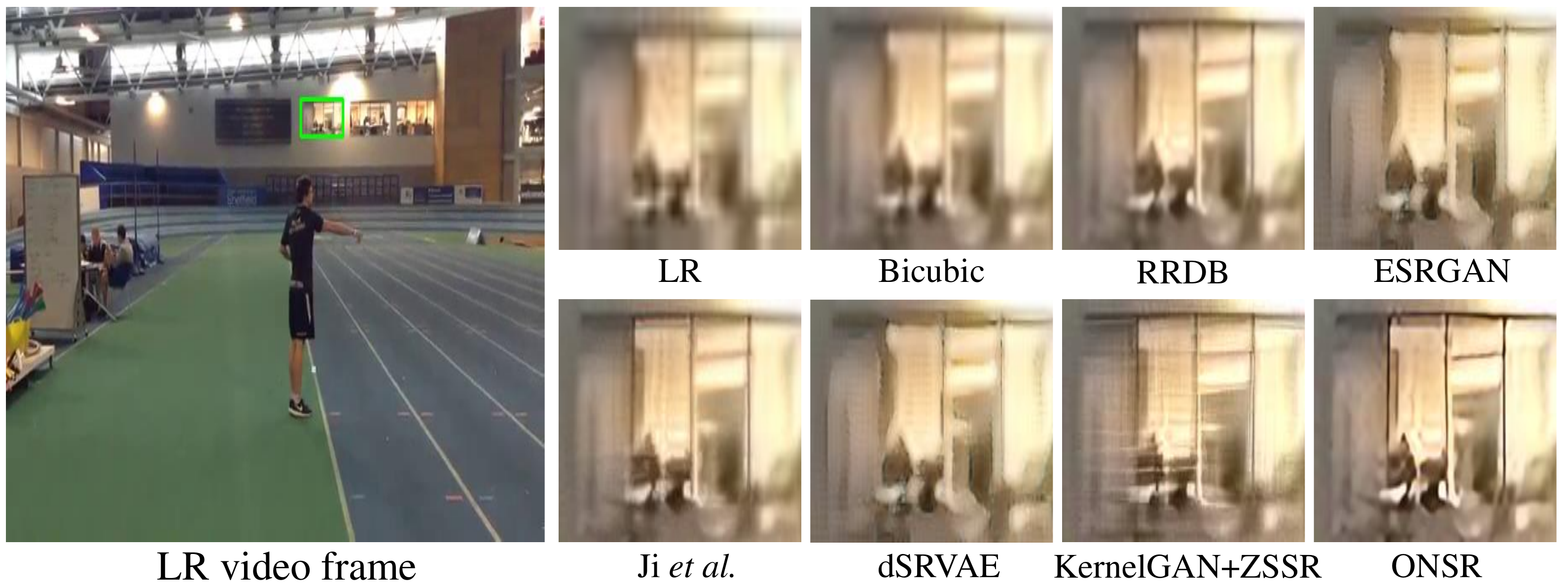}
	\caption{Visual comparison of model adaptation to  the real image ``\textit{Chip}'' and real-world video frames (from YouTube) for $4\times$ SR.}
	\label{f:ytb}
	\vspace{0mm}
\end{figure}

\subsection{Super-Resolution on Real-World Data}

To further demonstrate the effectiveness of ONSR in real scenarios, we conduct experiments on real images, which are more challenging due to the complicated and unknown degradations. Since there are no ground-truth HR images for calculating quantitative metrics, we only provide visual comparisons. As shown in Figure~\ref{f:ytb}, the letter ``X'' in the real image ``\textit{Chip}" restored by RRDB, RCAN, Ji \etal and ZSSR are blurry or have unpleasant artifacts. As a comparison, the super-resolved image of ONSR has more shaper edges and is more visually pleasing. We also apply these methods to video frames from YouTube \footnote{\url{https://www.youtube.com}}. As shown in Figure~\ref{f:ytb}, the generated $4\times$ SR frames from most methods are seriously blurred or contain numerous mosaics. While ONSR can produce visually promising images with clearer edges and fewer artifacts. This comparison further demonstrates the robustness of ONSR against various degradations in real scenarios.

\begin{figure}[h]
	\centering
	\includegraphics[width=\linewidth]{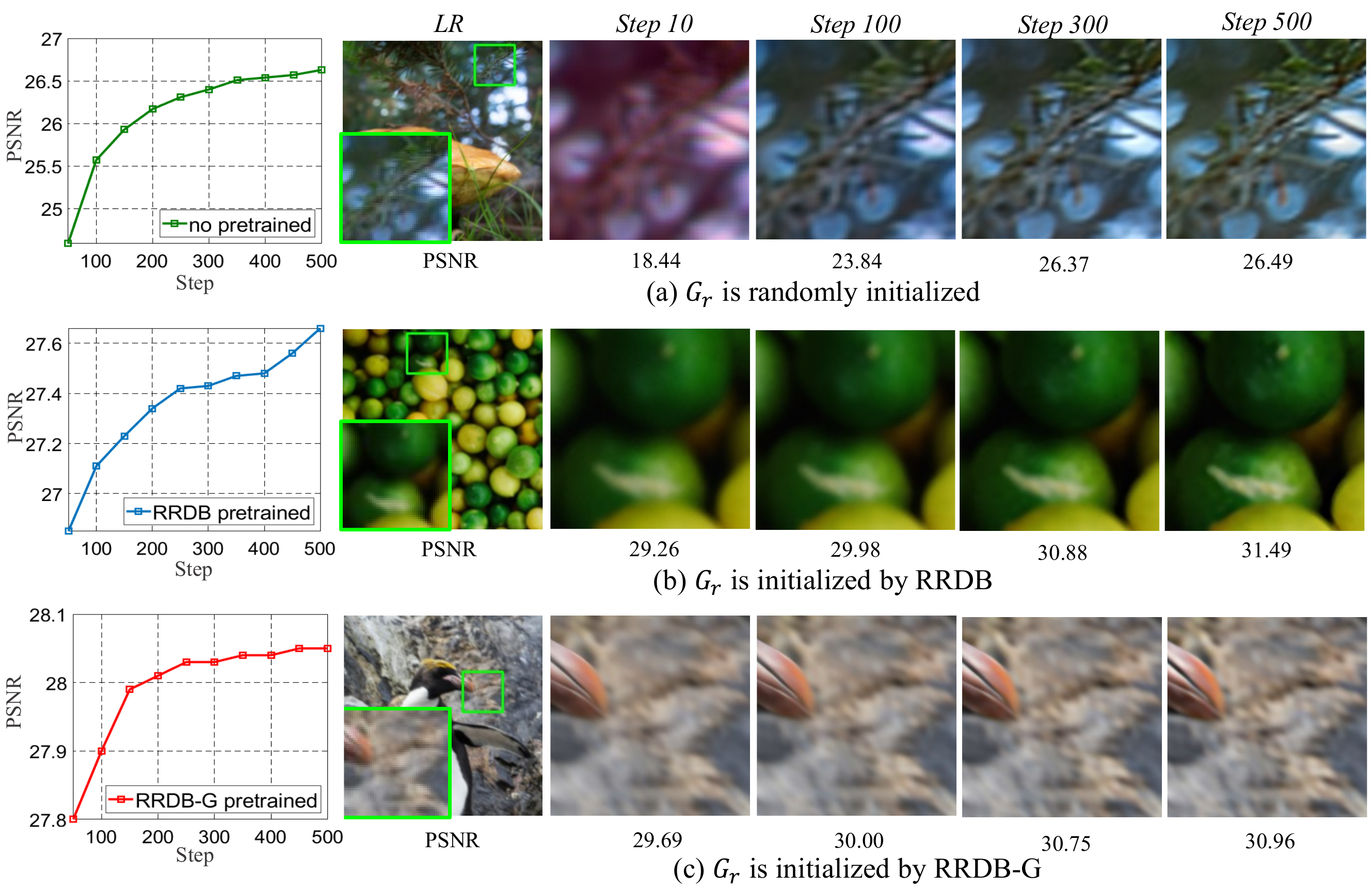}
	\caption{PSNR and visual results of ONSR with $G_r$ is initialized (a) randomly, (b) by model trained on data synthesized by bicubic downsampling, and (c) by model trained on data synthesized by multiple Gaussian kernels. }
	\label{f:pretrain}
\end{figure}

\begin{table}[H]
	\caption{Quantitative comparison between ONSR with other SotA blind SR methods. The $G_r$ in ONSR is initialized with RRDB-G, in which case our method is denoted as ONSR-G. Results on DIV2KRK are reported. Best and second best results are \textbf{highlighted} and \underline{underlined}}
	\label{t:sotag}
	\centering
	\setlength{\tabcolsep}{0.5cm}
	\resizebox{\linewidth}{!}{
		\begin{tabular}{lccccc}
			\toprule
			Method   & Scale                    & PSNR  $\uparrow$    & SSIM  $\uparrow$     & PI$\downarrow$       & LPIPS$\downarrow$    \\
			\midrule
			IKC \cite{ikc}&\multirow{4}{*}{$\times$2}
			& \underline{31.20} & \underline{0.8767}         & 5.1511             & 0.2350             \\
			DASR \cite{dasr}&
			& 30.72 & 0.8606         & 5.3947             & 0.2501             \\
			RRDB-G \cite{esrgan}&
			& 31.18             & 0.8763             & \underline{4.8995} & \underline{0.2213}         \\
			ONSR-G (Ours) &
			& \textbf{31.69}  & \textbf{0.8907}          & \textbf{4.6036}          & \textbf{0.1947}     \\ 
			\midrule
			KernelGAN~\cite{kernel_gan}  + USRNet~\cite{usr}& \multirow{5}{*}{$\times$4}
			& 24.32             & 0.6617            & 8.4425             & 0.5413             \\
			IKC \cite{ikc}&
			& 27.69             & 0.7657            & 6.9027             & 0.3863             \\
			DASR \cite{dasr}&
			& 27.48            & 0.7549           & 7.2024            & 0.4027            \\
			RRDB-G \cite{esrgan}&
			& \underline{27.73} & \underline{0.7660} & \underline{6.8767} & \underline{0.3834}  \\
			ONSR-G (Ours)&
			& \textbf{28.05}  & \textbf{0.7775}  & \textbf{6.7716}  & \textbf{0.3781}  \\
			\bottomrule
	\end{tabular}}
\end{table}

\subsection{Ablation Study}\label{sec:4.4}

\subsubsection{Study on the initialization of $G_r$}

As we have discussed above, the SR module  $G_r$ in our ONSR is usually initialized with pretrained model weights to reduce the updating steps. In this section, we experimentally investigate the influence of different initialization methods. Two initialization methods are firstly considered: a) random initialization, and b) model pretrained on data synthesized by bicubic downsampling. As shown in Figure~\ref{f:pretrain} (a) and (b), ONSR can converge well under both initialization, which demonstrates the good convergence of ONSR. In the meanwhile,  carefully initialized $G_r$ (case (b))  helps ONSR converge faster and has a better optimum point. It indicates that more powerful initialized $G_r$ may further improve the performance of ONSR.

\begin{figure}[t]
	\centering
	\includegraphics[width=0.9\linewidth]{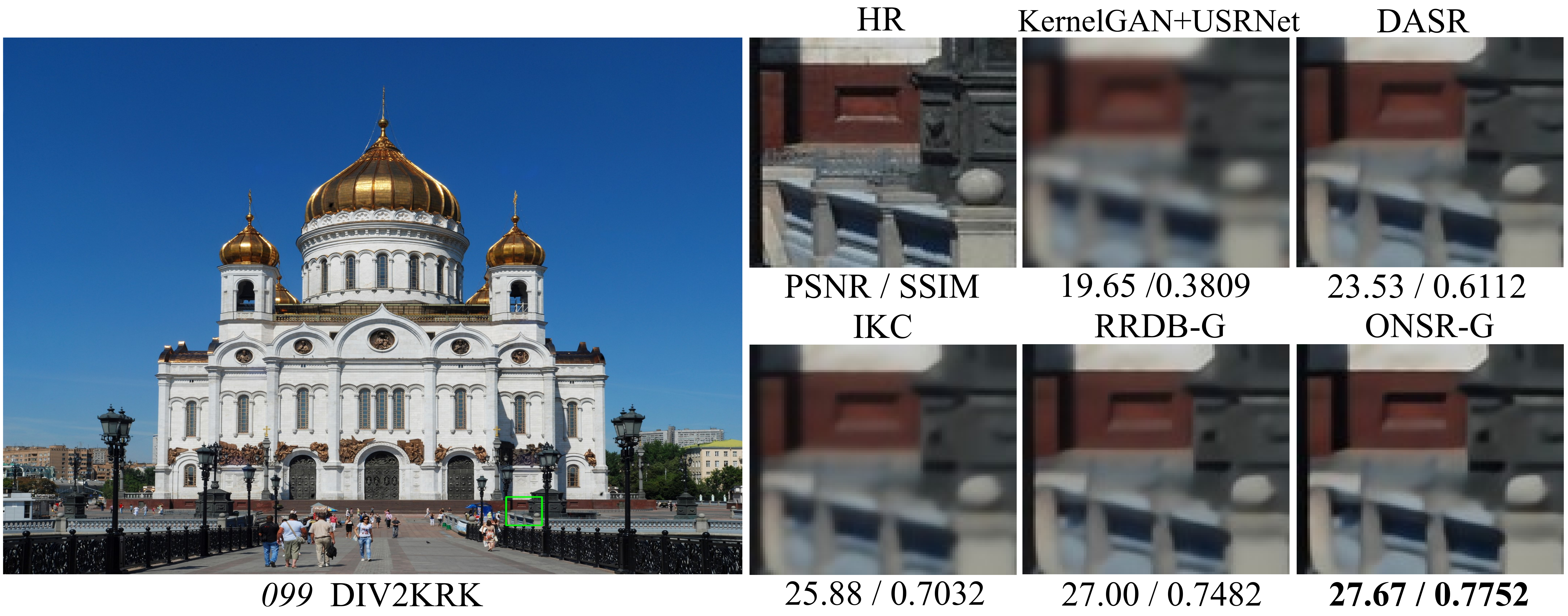}
	\includegraphics[width=0.9\linewidth]{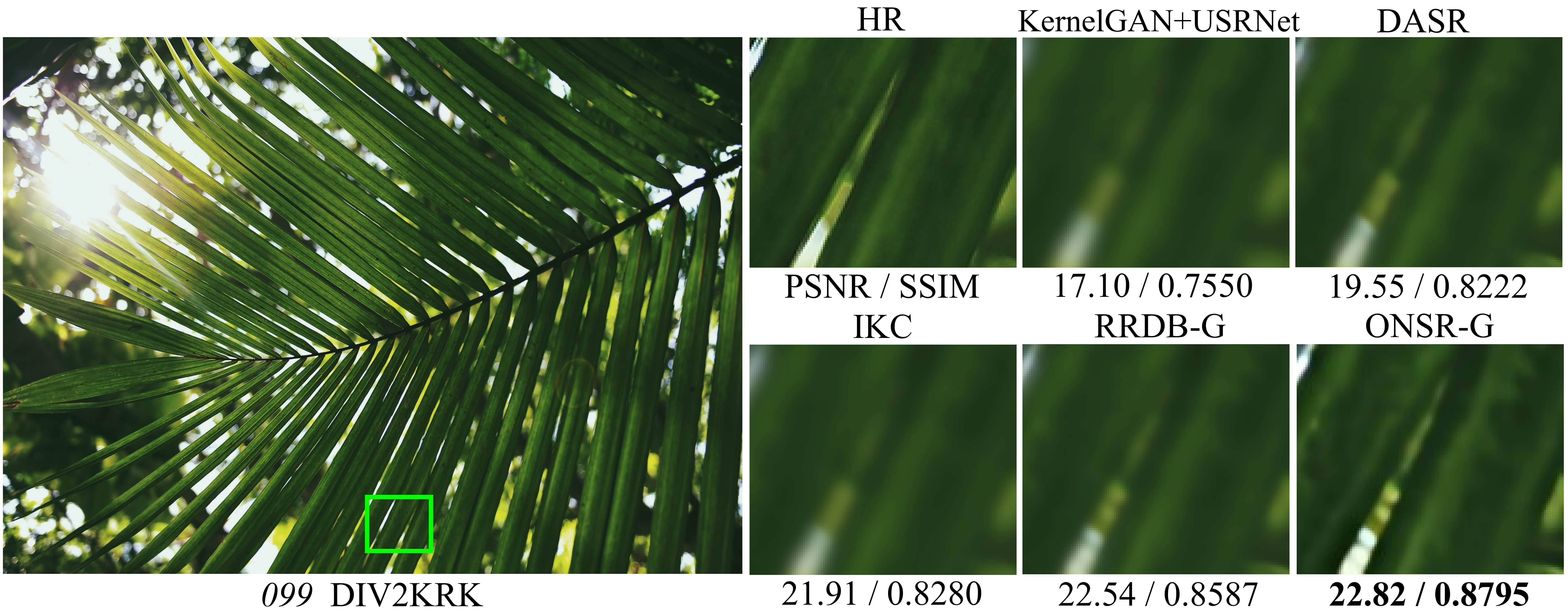} 
	\caption{Visual comparison of ONSR-G and other methods for $\times4$ SR (image \textit{026} and image \textit{099}). }
	\label{f:visual_gas}
\end{figure}

{ Recently,  some methods (such as IKC~\cite{ikc} and DASR~\cite{dasr}) are proposed to train the SR model with data synthesized by multiple Gaussian kernels to improve its robustness. This strategy has proved to be effective in training an accurate model for blind SR~\cite{xie2021finding}. To explore the performance of ONSR when $G_r$ is initialized with such a powerful model, we also train an RRDB on data synthesized by multiple isotropic Gaussian kernels, which is denoted as RRDB-G. We initialize $G_r$ with RRDB-G, in which case our method is denoted as ONSR-G. As shown in Figure~\ref{f:pretrain} (c), ONSR-G can still further improve the performance of the initialization model from $27.80$ dB to a better point of $28.05$ dB.  As shown in Table~\ref{t:sotag}, ONSR-G outperforms IKC by about $\mathbf{+0.4}$ dB on PSNR and $+\mathbf{0.012}$ on SSIM for scale factors $\times 2$ and $\times 4$ respectively. The visual comparisons are shown in Figure~\ref{f:visual_gas}. As one can see, the results produced by ONSR-G are clearer and more visually favorable. This comparison indicates that our method can easily enjoy the merits of other powerful SR methods by taking them as the initialization models.}

\begin{table}[h]
	\caption{The performance of ONSR with $G_r$ of different architectures. ON-$Arch$ denotes ONSR with the $G_r$ of architecture named $Arch$. Results are reported on DIV2KRK with scale factor  $\times4$. }
	\label{t:online}
	\centering
	\setlength{\tabcolsep}{1cm}{
		\resizebox{\linewidth}{!}{
			\begin{tabular}{lcccc}
				\toprule
				Method  & PSNR$\uparrow$  & SSIM$\uparrow$   & PI$\downarrow$ & LPIPS$\downarrow$  \\ \midrule
				RDN~\cite{rdn}    & 25.66 & 0.6935 & 8.5341 & 0.5411 \\
				ON-RDN  &\textbf{27.30} & \textbf{0.7498} & \textbf{7.4274} & \textbf{0.4377} \\ 
				\midrule
				RCAN~\cite{rcan}   & 24.75 & 0.6337 & 8.4560 & 0.5830 \\
				ON-RCAN & \textbf{27.58} & \textbf{0.7612} & \textbf{7.1290} & \textbf{0.4020} \\
				\bottomrule
			\end{tabular}
	}}
\end{table}

\subsubsection{Study on the architecture of $G_r$} 
In the experiments above, we only use RRDB as $G_r$ in our ONSR. In this subsection, we experimentally prove that the proposed method works well for $G_r$ of different architectures. Specifically, we perform experiments on $G_r$ of two other architectures, \ie RDN~\cite{rdn} and RCAN~\cite{rcan}. As shown in Table~\ref{t:online}, the original RDN and RCAN do not perform well on DIV2KRK. {This is because their original models are trained on data synthesized by bicubic downsampling and can not well generalize to test images in DIV2KRK. As a comparison, Our online updating strategy can largely improve the performance of both models (denoted as ON-RDN and ON-RCAN respectively). For example, as shown in~\ref{t:online}, the online updating strategy improves the PSNR results of  RDN and RCAN by $\mathbf{+1.64}$ dB and $\mathbf{+2.73}$~dB respectively. It indicates that the online updating strategy of ONSR works well for $G_r$ of different architectures.}

\begin{figure}[b]
	\centering
	\includegraphics[width=0.9\linewidth]{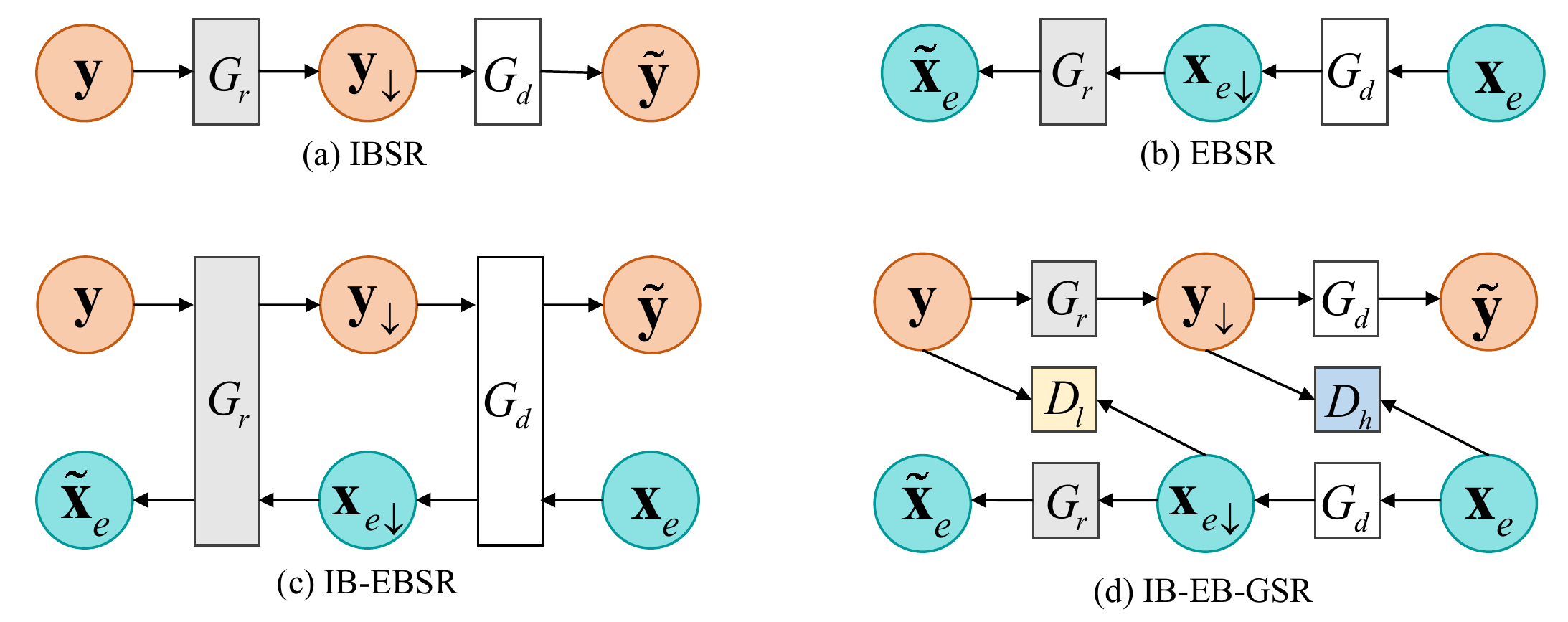}
	\caption{Details of different variants of ONSR.}
	\label{f:ablation}
\end{figure}

\subsubsection{Study on different modules in ONSR}\label{sec:4.4.3}
To explain the roles of different modules (\ie IB, EB and $D_l$) played in ONSR, we study four different variants of ONSR, which are denoted as \textit{IBSR}, \textit{EBSR}, \textit{IB-EBSR}, and \textit{IB-EB-GSR}. The details of these variants are shown in Figure~\ref{f:ablation} and explained below:

\textbf{IBSR.} IBSR only has an internal branch to exploit the internal properties of the test LR image for degradation estimation and SR reconstruction, which is optimized online.

\textbf{EBSR.} Contrary to IBSR, EBSR only has an external branch to capture general priors of external HR images, which is optimized offline. After offline training, we use the fixed module $G_r$ to test LR images. 

\textbf{IB-EBSR.} IB-EBSR has both internal branch and external branch but no GAN modules.

\textbf{IB-EB-GSR.} IB-EB-GSR has both $D_l$ and $D_h$ to explore the underlying distribution characteristics of the test LR and external HR images.

\begin{table}[H]
	\caption{Comparison of different variants of ONSR. Results are reported on DIV2KRK.}
	\label{t:bicycle}
	\centering
	\setlength{\tabcolsep}{0.6cm}{
		\resizebox{\linewidth}{!}{
			\begin{tabular}{lcllcll}
				\toprule
				Method    & \multicolumn{1}{l}{Scale} & PSNR  & SSIM   & Scale               & PSNR  & SSIM   \\ \midrule
				IBSR & \multirow{5}{*}{$\times$2}       & 28.05 & 0.8277 & \multirow{5}{*}{$\times$4} & 25.51 & 0.6976 \\
				EBSR &                           & 30.82 &  0.8806 &                     & 26.56 & 0.7249 \\
				IB-EBSR  &                           & 31.10 & 0.8850 &                     &27.60 & 0.7609 \\
				IB-EB-GSR &                           & 31.29 & 0.8859 &                     & 27.34 & 0.7507  \\ 
				ONSR &                           & \textbf{31.34} & \textbf{0.8866} &                     & \textbf{27.66} & \textbf{0.7620} \\ 
				\bottomrule
	\end{tabular}}}
\end{table}

The quantitative comparisons on DIV2KRK  are shown in Table~\ref{t:bicycle}. As one can see, IB-EBSR outperforms both IBSR and EBSR by a large margin. It indicates that both IB and EB are important for SR performance. The performance of IB-EBSR could be further improved if $D_l$ is introduced. It suggests that adversarial training can help $G_r$ to be better optimized. However, when $D_l$ and $D_h$ are both added in IB-EB-GSR, the performance is inferior to ONSR. In IB-EB-GSR, the initial SR results of  $G_r(\mathbf{y})$ are likely to have unpleasant artifacts or distortions. Besides, the external HR image $\mathbf{x}_e$ can not provide directly pixelwise supervision to $G_r(\mathbf{y})$. Therefore, the application of $D_h$ may affect the optimization of IB-EB-GSR and make it inferior to ONSR.

\subsubsection{Study on separate optimization}\label{exp_sep_opt}
\begin{figure}[h]
	\centering
	\includegraphics[width=\linewidth]{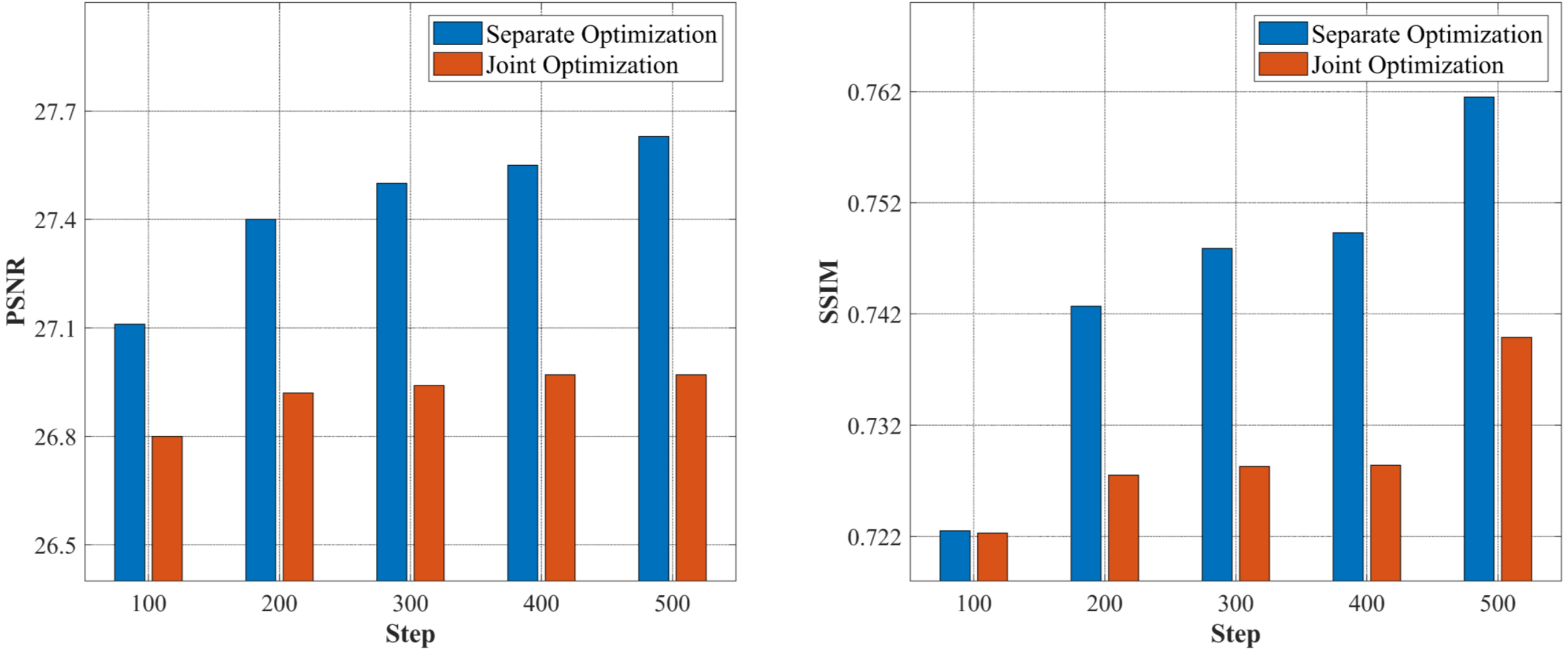}
	\caption{The average PSNR (left) and SSIM (right) of \textit{Joint Optimization} and \textit{Separate Optimization} for $4 \times$ SR in differente training steps.}
	\label{f:opt}
\end{figure}

As we have mentioned in Sec~\ref{sec:3.4}, we adopt separate optimization instead of the typically used joint optimization in our ONSR. In this section, we experimentally compare the two optimization strategies. In separate optimization, $G_d$ and $G_r$ are alternately optimized via the test LR image and external HR images respectively. While in joint optimization, both modules are optimized together. As shown in Table~\ref{t:opt}, separate optimization surpasses the joint optimization in all metrics for scale factors $\times2$ and $\times4$. We also compare the convergence of these two optimization strategies. We plot the PSNR and SSIM results of the two strategies every $100$ steps. As shown in Figure~\ref{f:opt},  the results of separate optimization are not only higher but also grow faster than that of joint optimization. It suggests that separate optimization could not only help the network converge faster, but also help it converge to a better optimum point. {It may be because that separate optimization divides the original highly complex problem into two simpler subproblems, and both of them are easier to be solved. Thus separate optimization could converge faster and reach a better optimum point}. This property of separate optimization allows us to make a trade-off between SR effectiveness and efficiency by setting different training iterations. 

\begin{table}[h]
	\caption{Comparisons between separate optimization and joint optimization. Results are reported on DIV2KRK.}
	\label{t:opt}
	\centering
	\setlength{\tabcolsep}{0.7cm}{
		\resizebox{\linewidth}{!}{
			\begin{tabular}{lccccc}
				\toprule
				Method                &            Scale            &  PSNR$\uparrow$  &  SSIM$\uparrow$   &  PI$\downarrow$   & LPIPS$\downarrow$ \\ 
				\midrule
				Joint Optimization 
				& \multirow{2}{*}{$\times$2} &     31.03      &     0.8827      &     4.8759      &    0.2212      \\
				Separate Optimization &                             & \textbf{31.34} & \textbf{0.8860} & \textbf{4.7952} & \textbf{0.2207} \\ 
				\midrule
				Joint Optimization    & \multirow{2}{*}{$\times$4} &     26.97      &     0.7399      &     7.5985      &     0.4445      \\
				Separate Optimization &                             & \textbf{27.66} & \textbf{0.7620} & \textbf{7.2298} & \textbf{0.4071} \\
				\bottomrule
			\end{tabular}
	}}
\end{table}

\subsubsection{Study on the External data}
In this section, we perform experiments to study the influence of the external HR data. Firstly, we investigate how the number of external HR images will influence SR performance.  we randomly sample 200, 400, 600, 700, and all 800 HR images from the training set of DIV2K as the external data and perform $4\times$ SR on DIV2KRK. As shown in Table~\ref{t:exhr}, the performance of ONSR is continuously improved as the number of HR images increases. It indicates that more external HR images could help ONSR learn better general priors. Secondly, we investigate the influence of different external HR datasets. We use another popular dataset Flickr2K that consists of 2650 HR images as the external dataset. As can be seen from Table~\ref{t:exhr}, the SR performance achieved with DIV2K is comparable to that with Flickr2K. Therefore, the SR performance tends to be stable when the number of external HR images is large enough, and  800 HR images from DIV2K could be sufficient for one test LR image.

\begin{table}[H]
	\caption{Performance of ONSR when different number of external HR images are used. Results are reported on DIV2KRK.}
	\vspace{0.01\linewidth}
	\label{t:exhr}
	\centering
	\setlength{\tabcolsep}{0.4cm}
	\resizebox{\linewidth}{!}{
		\begin{tabular}{ccccccc}
			\toprule
			\# External HR & 200          & 400          & 600          & 700          & 800  (DIV2K)          & 2650 (Flickr2K) \\
			\midrule
			PSNR    & 26.93 & 26.97 & 27.58 & 27.60 & 27.66 &  27.65    \\ 
			SSIM    & 0.7390 & 0.7399 & 0.7606 & 0.7610 & 0.7620 &  0.7629    \\ 
			\bottomrule
	\end{tabular}}
\end{table}

\subsection{Non-Blind Setting}
In this subsection, we explore the upper boundary of ONSR in a non-blind setting. We replace $G_d$ in ONSR with the ground-truth blur kernel (denoted as ONSR-NonBlind). Following the setting in~\cite{usr}, we evaluate our methods on \textit{BSD68}~\cite{bsd100}, and 12 representative and diverse blur kernels are used to synthesize test LR images, including 4 isotropic Gaussian kernels with different widths, 4 anisotropic Gaussian kernels from \cite{srmd}, and 4 motion blur kernels from \cite{Boracchi2012ModelingTP,Levin2009UnderstandingAE}. The quantitative results are shown in Table~\ref{t:non-blind}. {As one can see, under the non-blind setting, ONSR achieves the best performance among the reference methods. We need to note that USRNet~\cite{usr} is a SotA non-blind method, while ONSR still outperforms it by a large margin on all 12 kernels. For example, under the first blur kernel, ONSR outperforms USRNet by $+\mathbf{1.11}$ dB and $+\mathbf{1.21}$ dB for scale factors $\times2$ and $\times4$ respectively. It demonstrates that the online updating strategy of ONSR also works well in a non-blind setting. }

\begin{table}[H]
	\centering
	\caption{Average PSNR results of non-blind setting for $4\times$ SR. Best and second best results are \textbf{highlighted} and \underline{underlined}.}\label{t:non-blind}
	\setlength{\tabcolsep}{0.1mm}
	\resizebox{\linewidth}{!}{
		\begin{tabular}{lccccccccccccc}
			\toprule
			Method & Scale               & \begin{minipage}{0.1\linewidth}
				\centering
				\includegraphics[width=0.7\linewidth]{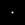}
			\end{minipage}
			
			& 
			\begin{minipage}{0.1\linewidth}
				\centering
				\includegraphics[width=0.7\linewidth]{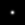}
			\end{minipage}
			& 
			\begin{minipage}{0.1\linewidth}
				\centering
				\includegraphics[width=0.7\linewidth]{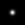}
			\end{minipage}
			& 
			\begin{minipage}{0.1\linewidth}
				\centering
				\includegraphics[width=0.7\linewidth]{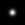}
			\end{minipage}
			& 
			\begin{minipage}{0.1\linewidth}
				\centering
				\includegraphics[width=0.7\linewidth]{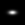}
			\end{minipage}
			& 
			\begin{minipage}{0.1\linewidth}
				\centering
				\includegraphics[width=0.7\linewidth]{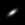}
			\end{minipage}
			&
			\begin{minipage}{0.1\linewidth}
				\centering
				\includegraphics[width=0.7\linewidth]{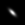}
			\end{minipage}
			& 
			\begin{minipage}{0.1\linewidth}
				\centering
				\includegraphics[width=0.7\linewidth]{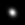}
			\end{minipage}
			& 
			\begin{minipage}{0.1\linewidth}
				\centering
				\includegraphics[width=0.7\linewidth]{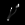}
			\end{minipage}
			& 
			\begin{minipage}{0.1\linewidth}
				\centering
				\includegraphics[width=0.7\linewidth]{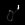}
			\end{minipage}
			& 
			\begin{minipage}{0.1\linewidth}
				\centering
				\includegraphics[width=0.7\linewidth]{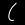}
			\end{minipage}
			& 
			\begin{minipage}{0.1\linewidth}
				\centering
				\includegraphics[width=0.7\linewidth]{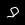}
			\end{minipage}   \\
			\midrule
			EDSR ~\cite{edsr}   &   \multirow{6}{*}{$\times$2}                   &   25.54
			&   27.82   &    20.59   &  21.34   &   27.66    &    27.28   &   26.90    &  26.07     &   27.14   &   26.96   &   19.72   &   19.86   \\
			RCAN ~\cite{rcan}   &                     &   29.48   &   26.76   &    25.31 &   24.37    &   24.38   &    24.10   &   24.25    &  23.63     &   20.31   &   20.45    &   20.57    &   22.04   \\
			ZSSR~\cite{zssr}   &    & 29.44 & 29.48 & 28.57 & 27.42 & 27.15 & 26.81 & 27.09 & 26.25 & 14.22 & 14.22 & 16.02 & 19.39 \\
			IRCNN~\cite{ircnn}  &                     & 29.60 & 30.16 & 29.50 & 28.37 & 28.07 & 27.95 & 28.21 & 27.19 & 28.58 & 26.79 & 29.02 & 28.96 \\
			USRNet~\cite{usr} &                     & \underline{30.55} & \underline{30.96} & \underline{30.56} & \underline{29.49} & \underline{29.13} & \underline{29.12} & \underline{29.28} & \underline{28.28} & \underline{30.90} & \underline{30.65} & \underline{30.60} & \underline{30.75} \\
			ONSR-NonBlind   &                     & \textbf{31.66} & \textbf{31.98} & \textbf{31.40} & \textbf{30.17} & \textbf{29.76} & \textbf{29.63} &\textbf{29.86} & \textbf{28.87} & \textbf{30.93} & \textbf{30.78} & \textbf{30.80} & \textbf{31.12} \\
			\midrule
			EDSR ~\cite{edsr}   &   \multirow{6}{*}{$\times$4}    &   21.45    &   22.73  &    21.60   &   20.62  &   23.16    &    23.66   &   23.16    &  23.00    &   24.00    &   23.78    &   19.79    &   19.67   \\
			RCAN ~\cite{rcan}   &                     &   22.68    &   25.31   &    25.59  &   24.63   &   24.37    &    24.23   &   24.43    &  23.74    &   20.06   &   20.05   &   20.33    &   21.47    \\
			ZSSR~\cite{zssr}   &     & 23.50 & 24.33 & 24.56 & 24.65 & 24.52 & 24.20 & 24.56 & 24.55 & 16.94 & 16.43 & 18.01 & 20.68 \\
			IRCNN~\cite{ircnn}  &                     & 23.99 & 25.01 & 25.32 & 25.45 & 25.36 & 25.26 & 25.34 & 25.47 & 24.69 & 24.39 & 24.44 & 24.57 \\
			USRNet~\cite{usr} &                     & \underline{25.30} & \underline{25.96} & \underline{26.18} & \underline{26.29} & \underline{26.20} & \underline{26.15} & \underline{26.17} & \underline{26.30} & \underline{25.91} & \underline{25.57} & \underline{25.76} & \underline{25.70} \\
			ONSR-NonBlind   &                     &\textbf{26.51} & \textbf{27.24} & \textbf{27.50} &\textbf{27.57} & \textbf{27.43} & \textbf{27.30} & \textbf{27.36} & \textbf{27.51} &\textbf{26.17} &\textbf{26.17} & \textbf{26.21} &\textbf{26.30} \\
			\bottomrule
	\end{tabular}}
\end{table}

\subsection{Speed Comparison }
{We test the speed of different blind SR methods to compare their overall performance in terms of effectiveness and efficiency. We evaluate their average running time on DIV2KRK for $\times2$ on the same machine with an NVIDIA 2080Ti GPU. Since IKC~\cite{ikc} and DASR~\cite{dasr} are included in the referring methods, we report the performance of  ONSR-G for fair comparisons. We need to note that the running time of ONSR-G is closely related to its online updating steps. In previous experiments, the steps are set as 500 for best performance. However, as we have discussed in Sec~\ref{exp_sep_opt}, due to the good convergence of ONSR-G, we can set fewer steps to make a balance between accuracy and speed. Thus, we also test the performance of ONSR-G when steps are 5, 10,  and 100 for comparisons. We use ONSR-G-$S$  to denote ONSR-G with updating steps as $S$.  As shown in Table~\ref{t:times},  ONSR-G-5 outperforms KernelGAN + ZSSR~\cite{kernel_gan} and Cornillere \etal by $+\mathbf{4.39}$ dB and $+\mathbf{1.73}$ dB while with $\mathbf{174}$ times and $21$ times faster speed respectively. When compared with IKC, ONSR-G-10 also achieves a similar PSNR result with a comparable speed. DASR is much faster than the other methods, but its PNSR result is suboptimal. On the whole, ONSR-G can achieve competing performance with the most recent blind SR methods. Moreover, the good convergence of ONSR allows us to easily adjust its running time according to different scenarios. }

\begin{table}[h]
	\caption{Average inference time comparison on DIV2KRK for $\times2$ SR.  ONSR-G-$S$ denotes ONSR-G with updating steps as $S$. }
	\label{t:times}
	\centering
	\setlength{\tabcolsep}{15mm}{
		\resizebox{\linewidth}{!}{
			\begin{tabular}{lcc}
				\toprule
				Method         & Speed(s/image) & PSNR(dB) \\ 
				\midrule
				KernelGAN + ZSSR~\cite{kernel_gan} & 1127.84  & 26.76\\
				Cornillere \etal~\cite{cornillere2019blind} & 135.51  & 29.42\\
				IKC~\cite{ikc}    & 7.14  & 31.20\\
				DASR~\cite{dasr}	& 0.24  & 30.72\\
				
				ONSR-G-5 (Ours)     & 6.47   & 31.15  \\
				ONSR-G-10 (Ours)     & 8.00   & 31.21    \\ 
				ONSR-G-100 (Ours)     & 75.29   & 31.61    \\
				\bottomrule
			\end{tabular}
	}}
\end{table}

\section{Conclusion and Future Work}
In this paper, we argue that most nowadays SR methods are not image-specific. Towards the limitation, we propose an online SR (ONSR) method, which could customize a specific model for each test image. In detail, we design two branches, namely internal branch (IB) and external branch (EB).
IB could learn the specific degradation of the test image, and EB could learn to super resolve images that are degraded by the learned degradation. IB involves only the LR image, while EB uses external HR images. In this way, ONSR could leverage the benefits of both inherent information of the test LR image and general priors from external HR images. Extensive experiments on both synthetic and real-world images prove the superiority of ONSR in the blind SR problem. These results indicate that customizing a model for each test image is more practical in real applications than training a general model for all LR images. Moreover, the speed of  ONSR may be further improved by designing more lightweight modules for faster inference or elaborating the training strategy to accelerate convergence. Faster speed can help it to be more practical when processing large amounts of test images, such as videos of low resolution, which is also the focus of our future work. 

\bibliographystyle{elsarticle-num}

\end{document}